\documentclass[runningheads]{llncs}

\usepackage{eccv}

\usepackage{eccvabbrv}

\usepackage{graphicx}
\usepackage{booktabs}
\definecolor{cvprblue}{rgb}{0.21,0.49,0.74}
\usepackage[accsupp]{axessibility}  % Improves PDF readability for those with disabilities.

\usepackage[pagebackref,breaklinks,colorlinks,citecolor=eccvblue]{hyperref}
\usepackage{amsmath,amsfonts,bm,amssymb,mathtools}
\usepackage{color,xcolor,xspace}
\usepackage{booktabs}
\usepackage{thm-restate}
\usepackage{bbding}
\usepackage{pifont}
\usepackage{wasysym}

\newcommand{\bb}[1]{{\mathbb{#1}}}
\newcommand{\norm}[1]{{\lVert {#1} \rVert}}
\newcommand{\diff}{\mathrm{d}}

\DeclareMathOperator{\attention}{Attention}

\DeclareMathOperator{\softmax}{Softmax}
\def\eqref#1{Eq.(\ref{#1})}
\def\1{\bm{1}}

\def\rvz{{\mathbf{z}}}

\def\mI{{\bm{I}}}

\DeclareMathAlphabet{\mathsfit}{\encodingdefault}{\sfdefault}{m}{sl}
\SetMathAlphabet{\mathsfit}{bold}{\encodingdefault}{\sfdefault}{bx}{n}

\def\gN{{\mathcal{N}}}

\usepackage{url}

\usepackage[utf8]{inputenc} % allow utf-8 input
\usepackage[T1]{fontenc}    % use 8-bit T1 fonts
\usepackage{hyperref}       % hyperlinks
\usepackage{url}            % simple URL typesetting
\usepackage{booktabs}       % professional-quality tables
\usepackage{amsfonts}       % blackboard math symbols
\usepackage{nicefrac}       % compact symbols for 1/2, etc.
\usepackage{microtype}      % microtypography
\usepackage{xcolor}         % colors

\usepackage[utf8]{inputenc} 
\usepackage[T1]{fontenc}    
\usepackage{hyperref}      
\usepackage{url}           
\usepackage{booktabs}       
\usepackage{amsmath}       
\usepackage{nicefrac}     
\usepackage{microtype}      
\usepackage{multirow}      
\usepackage{graphicx}
\usepackage{graphics}
\usepackage{tabularx}
\usepackage{makecell}
\usepackage{caption}
\usepackage{wrapfig}
\usepackage{enumitem}
\usepackage{subcaption}
\usepackage{amsfonts}      
\usepackage{nicefrac}

\usepackage{enumitem}
\usepackage{algorithmic}
\usepackage{algorithm}
\usepackage{verbatim}
\usepackage{listings}
\usepackage{wrapfig}
\usepackage{mathtools}
\definecolor{codeblue}{rgb}{0.25,0.5,0.25}
\definecolor{codekw}{rgb}{0.85, 0.18, 0.50}
\usepackage{subcaption}
\usepackage{tikz}
\usepackage{comment}

\usepackage{color, colortbl}
\definecolor{Gray}{gray}{0.93}
\definecolor{orange}{rgb}{0.9,0.5,0}

\usepackage{hyperref} 
 \hypersetup{ 
     colorlinks=true, 
     linkcolor=red, 
     }

\usepackage[margin=4pt,font=small,labelfont=bf,labelsep=endash,tableposition=bottom]{caption}
\newcolumntype{H}{>{\setbox0=\hbox\bgroup}c<{\egroup}@{}}

\definecolor{mycolor}{HTML}{5a860e}

\usepackage{orcidlink}

\begin{document}

% ---------------------------------------------------------------
% TODO REVIEW: Replace with your title
\title{\textcolor{mycolor}{DiffiT}: \textcolor{mycolor}{Diff}usion V\textcolor{mycolor}{i}sion \textcolor{mycolor}{T}ransformers for Image Generation}

% TODO REVIEW: If the paper title is too long for the running head, you can set
% an abbreviated paper title here. If not, comment out.
\titlerunning{Diffusion Vision Transformers for Image Generation}

% TODO FINAL: Replace with your author list. 
% Include the authors' OCRID for the camera-ready version, if at all possible.
\author{Ali Hatamizadeh, Jiaming Song, Guilin Liu, Jan Kautz, Arash Vahdat}

% TODO FINAL: Replace with an abbreviated list of authors.
\authorrunning{Hatamizadeh et al.}
% First names are abbreviated in the running head.
% If there are more than two authors, 'et al.' is used.

% TODO FINAL: Replace with your institution list.
\institute{NVIDIA \\
\email{{\{ahatamizadeh, jiamings, guilinl, jkautz, avahdat\}@nvidia.com}\\
\url{https://github.com/NVlabs/DiffiT}}
}
\maketitle

\renewcommand\twocolumn[1][]{#1}%
\begin{center}
    \centering
    \captionsetup{type=figure}
    \includegraphics[width=\linewidth]{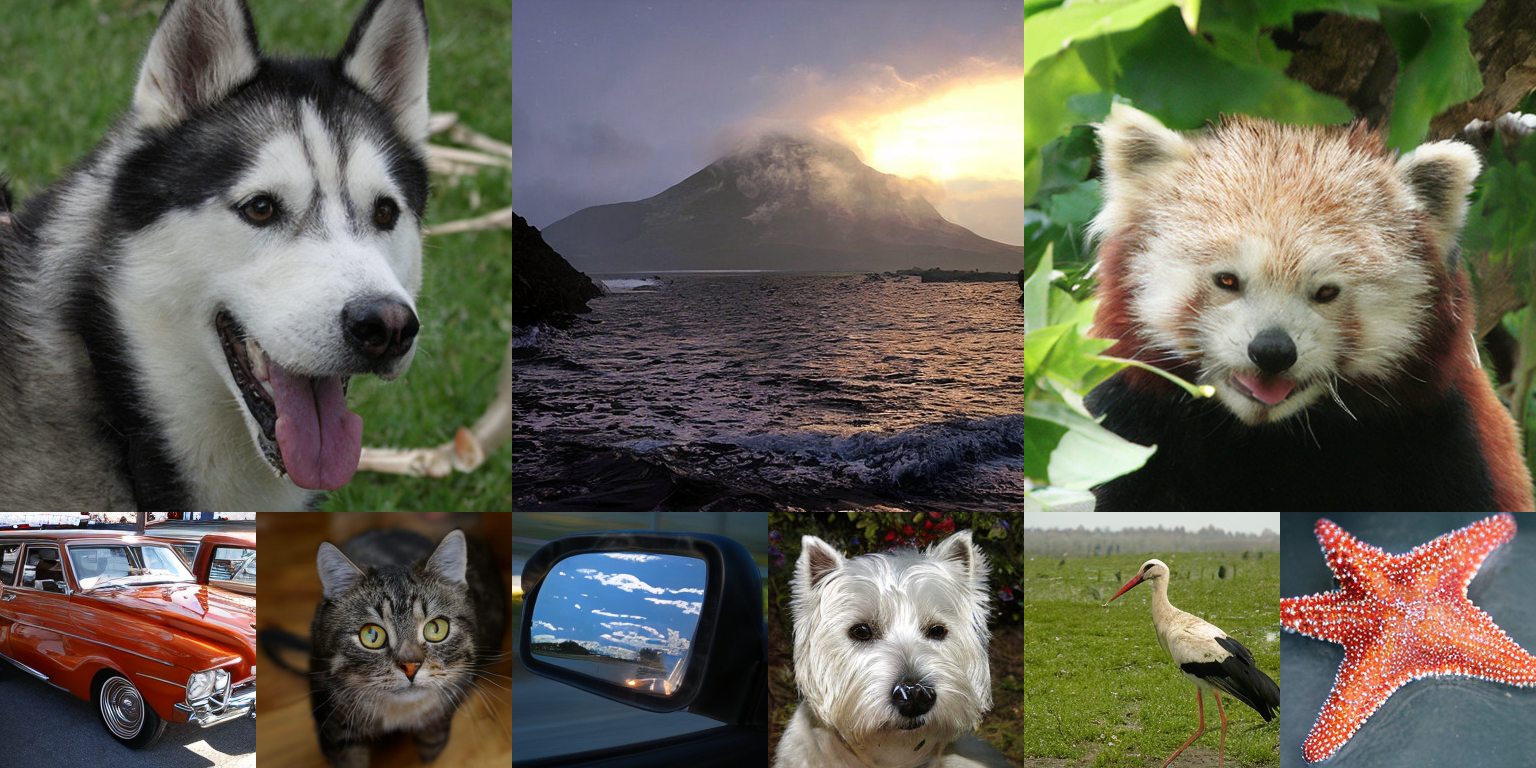}
    \vspace{-4mm}
    \captionof{figure}{Uncurated generated images by latent DiffiT on ImageNet~\cite{deng2009imagenet} dataset.}\label{fig:teaser}
\end{center}%

\begin{abstract}
  Diffusion models with their powerful expressivity and high sample quality have achieved State-Of-The-Art (SOTA) performance in the generative domain. The pioneering Vision Transformer (ViT) has also demonstrated strong modeling capabilities and scalability, especially for recognition tasks. In this paper, we study the effectiveness of ViTs in diffusion-based generative learning and propose a new model denoted as Diffusion Vision Transformers (DiffiT). Specifically, we propose a methodology for finegrained control of the denoising process and introduce the Time-dependant Multihead Self Attention (TMSA) mechanism. DiffiT is surprisingly effective in generating high-fidelity images with significantly better parameter efficiency. We also propose \emph{latent} and \emph{image} space DiffiT models and show SOTA performance on a variety of class-conditional and unconditional synthesis tasks at different resolutions. The Latent DiffiT model achieves a new SOTA FID score of \textbf{1.73} on \textbf{ImageNet-256} dataset while having $\mathbf{19.85\%}$, $\mathbf{16.88\%}$ less parameters than other Transformer-based diffusion models such as MDT and DiT, respectively.
\end{abstract}

%%%%%%%%% BODY TEXT
\section{Introduction}
\label{sec:intro}

Diffusion models~\cite{sohl-dickstein2015deep,ho2020denoising,song2020score} have revolutionized the domain of generative learning, with successful frameworks in the front line such as DALL$\cdot$E 3~\cite{ramesh2022hierarchical}, Imagen~\cite{ho2022imagen} and Stable diffusion~~\cite{rombach2022high,stablediffusion2022} and achieving state-of-the-art (SOTA) performance in various tasks. They have enabled generating diverse complex scenes in high fidelity which were once considered out of reach for prior models. Specifically, synthesis in diffusion models is formulated as an iterative process in which random image-shaped Gaussian noise is denoised gradually towards realistic samples~\cite{sohl-dickstein2015deep,ho2020denoising,song2020score}. The core building block in this process is a \textit{denoising autoencoder network} that takes a noisy image and predicts the denoising direction, equivalent to \textit{the score function}~\cite{vincent2011a, Hyvarinen05Score}. This network, which is shared across different time steps of the denoising process, is often a variant of Convolutional Neural Network (CNN)-based U-Net~\cite{ronneberger2015UNet, ho2020denoising}. However, with a lack of standard design pattern, several architecture variants~\cite{song2019generative, nichol2021improved, dhariwal2021diffusion} have been proposed for the denoising network. 

% Although the self-attention layers have shown to be important for capturing long-range spatial dependencies, yet there exists a lack of standard design patterns on how to incorporate them. 
%\texttt{Sora}~\cite{openai2024sora}

Vision Transformers (ViTs)~\cite{dosovitskiy2020image} have demonstrated SOTA performance for various recognition tasks and offer compelling advantages such as long-range dependency modeling and scalability. Recently, a number of efforts such as Diffusion Transformers (DiT)~\cite{peebles2022scalable} and Masked Diffusion Transformer (MDT)~\cite{gao2023masked} \begin{wrapfigure}{r}{0.58\textwidth}
\vspace{-7.5mm}
  \begin{center}
\includegraphics[width=1.0\linewidth,clip=True]{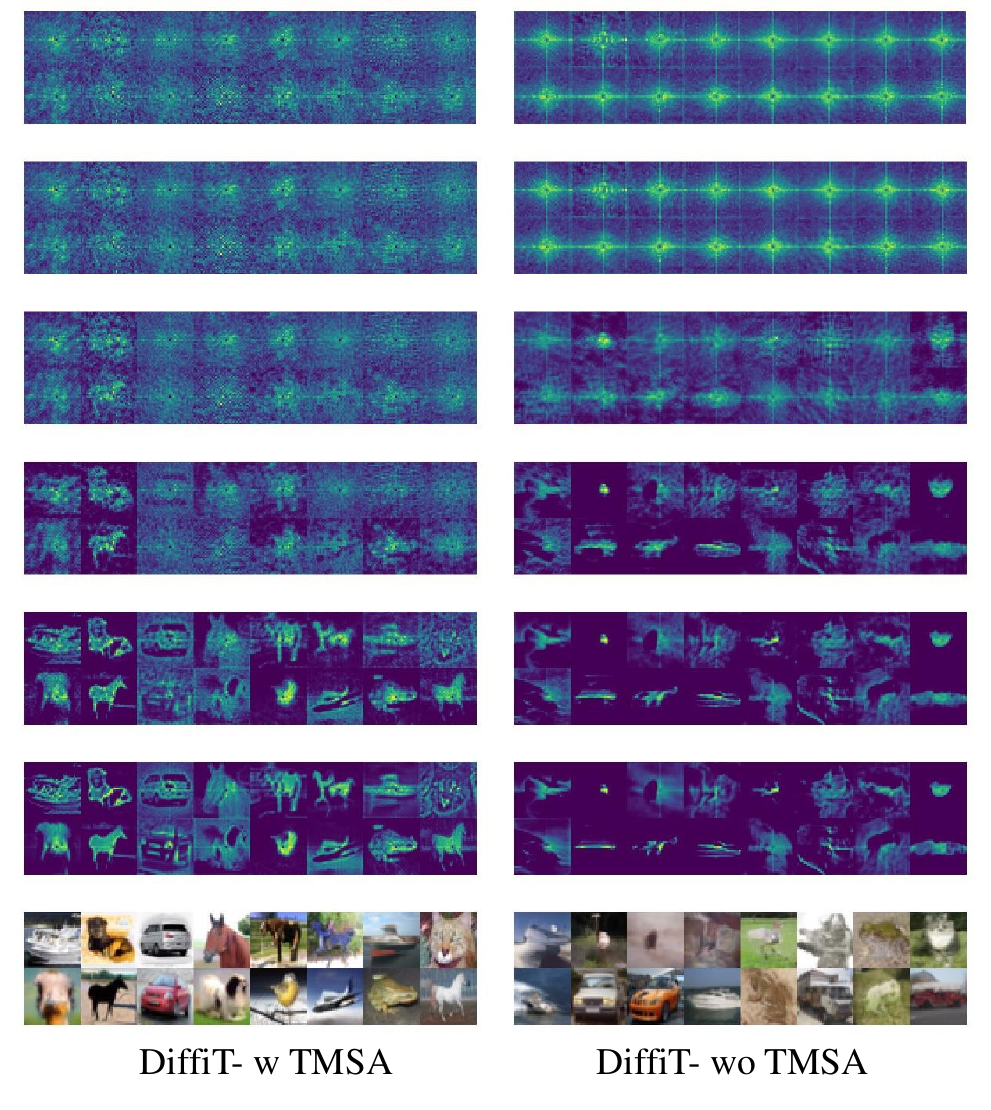}
  \end{center}
\caption{Side-by-side qualitative comparison of attention maps during the denoising process for models with and without TMSA. The denoising process starts from the top row in each column. }\label{fig:self-attn-evol}
\vspace{-5.5mm}
\end{wrapfigure} have proposed to leverage the strong modeling capability and scalability of ViTs for diffusion-based image generation. In DiT and MDT, Adaptive LayerNorm (AdaLN)~\cite{perez2018film} is used for input noise conditioning. However, this scheme significantly increases the number of parameters and does not effectively model the unique temporal dynamics of the denoising process~\cite{Kreis2022Tutorial, choi2022perception}. Specifically, in the beginning of denoising, the high-frequency content of the image is completely perturbed as the denoising network primarily focuses on predicting the low-frequency content. Towards the end of denoising, in which most of the image structure is generated, the network tends to focus on predicting high-frequency details. The conditioning in DiT is realized by modulating the input with channel-wise scale and shift parameters predicted by adaLN layers. However, this mechanism cannot optimally capture the dynamics of the denoising process since it does not effectively model the joint spatial and temporal dependencies. In this work, we introduce the Time-dependant Multihead Self-Attention (TMSA) mechanism which allows for fine-grained control over spatial and temporal dependencies and their interaction during the denoising process. Specifically, our TMSA proposes to integrate the temporal component into the self-attention where the key, query, and value weights are adapted per time step during denoising. This allows the denoising network to dynamically change its attention mechanism in different stages by considering both spatial and temporal components and their correspondence. In Fig.~\ref{fig:self-attn-evol}, we visualize attention maps from a token at the center of a feature map to all surrounding tokens during the sampling trajectory of a models that are trained on CIFAR10~\cite{krizhevsky2009learning} dataset. The DiffiT model with TMSA has a better image generation quality and its attention maps demonstrate a progressive localization towards detailed salient features. However, the model without TMSA is not capable of recovering such details. 

In addition, employing TMSA significantly improves the parameter efficiency as it only learns three temporal components for query, key and value in each block. In comparison, AdaLN requires learning the shift, scale and gate parameters for self-attention as well as MLP (\textit{i.e.} six components per Transformer block). We also extend TMSA to a window-based scheme without cross-communication among the local regions. This design is surprisingly effective and decreases the computational cost of self-attention by reducing the token sequence length.   

Using TMSA as a core building block, we introduce a novel ViT-based diffusion model, called DiffiT (pronounced \textit{di-feet}), for image generation in latent and image space. DiffiT achieves a new SOTA performance in terms of FID score using ImageNet-256~\cite{deng2009imagenet} dataset (see Fig.\ref{fig:teaser}) with $\mathbf{19.85\%}$, $\mathbf{16.88\%}$ less parameters than MDT and DiT models, respectively. DiffiT also achieves SOTA performance for image space generation tasks on FFHQ-64~\cite{karras2019style} and CIFAR10~\cite{krizhevsky2009learning} datasets.

The following summarizes our contributions in this work:
\begin{itemize}[noitemsep,nosep]
\item We introduce TMSA which is a novel time-dependent self-attention mechanism and is specifically tailored to capture both temporal and spatial dependencies as well as their interaction. Our proposed time-dependent self-attention dynamically adapts its behavior over sampling time steps.

\item We introduce a new ViT-based diffusion model, denoted as DiffiT, which unifies the design patterns of denoising networks and can be used in a variety of image generation tasks in the latent and image space.

\item We demonstrate that DiffiT can achieve SOTA performance on a variety of datasets for both conditional and unconditional generation tasks in the latent and image space. The latent DiffiT model achieves a new FID score of 1.73 with significantly less number of parameters than competing approaches.  
\end{itemize}

% The following summarizes our contributions in this work:

% \begin{itemize}[noitemsep,nosep]

% \end{itemize}

\section{Related Work}
\label{sec:related}
\paragraph{Diffusion Image Generation} Diffusion models~\cite{sohl-dickstein2015deep,ho2020denoising,song2020score} have driven significant advances in various domains, such as text-to-image generation~\cite{ramesh2022hierarchical,saharia2022photorealistic,balaji2022ediffi}, natural language processing~\cite{li2022diffusion}, text-to-speech synthesis~\cite{kong2020diffwave}, 3D point cloud generation~\cite{zeng2022lion,ye2022first,zhou20213d}, time series modeling~\cite{tashiro2021csdi}, molecular conformal generation~\cite{xu2022geodiff}, and machine learning security~\cite{nie2022DiffPure}. These models synthesize samples via an iterative denoising process and thus are also known in the community as noise-conditioned score networks. Since its initial success on small-scale datasets like CIFAR-10~\cite{ho2020denoising}, diffusion models have been gaining popularity compared to other existing families of generative models. Compared with variational autoencoders~\cite{kingma2013auto}, diffusion models divide the synthesis procedure into small parts that are easier to optimize, and have better coverage of the latent space~\cite{aneja2021contrastive,sinha2021d2c,vahdat2021score}; compared with generative adversarial networks~\cite{goodfellow2014generative}, diffusion models have better training stability and are much easier to invert~\cite{song2020denoising,gal2022image}. Diffusion models are also well-suited for image restoration, editing and re-synthesis tasks with minimal modifications to the existing architecture~\cite{meng2021sdedit,ruiz2022dreambooth,gal2022image,kawar2022imagic,avrahami2022blended,avrahami2022blendedlatent,couairon2022diffedit,kawar2022denoising,kawar2022jpeg,valevski2022unitune}, making it well-suited for various downstream applications. 

\paragraph{Transformers in Generative Modeling}
Transformer-based models have achieved competitive performance in different generative learning models in the visual domain~\cite{chen2020generative,zhang2022styleswin,ding2022cogview2,zhang2021ernie,hong2022cogvideo,ding2021cogview}. A number of transformer-based architectures have emerged for GANs~\cite{li2021can,xu2021stransgan,zhao2021improved,lee2021vitgan}. TransGAN~\cite{jiang2021transgan} proposed to use a pure transformer-based generator and discriminator architecture for pixel-wise image generation. Gansformer~\cite{hudson2021generative} introduced a bipartite transformer that encourages the similarity between latent and image features. Styleformer~\cite{park2022styleformer} uses Linformers~\cite{wang2020linformer} to scale the synthesis to higher resolution images. Recently, a number of efforts~\cite{luhman2022improving,bao2022all,peebles2022scalable,gao2023masked} have leveraged Transformer-based architectures for diffusion models and achieved competitive performance. In particular, DiT~\cite{peebles2022scalable} proposed a latent diffusion model in which the regular U-Net backbone is replaced with a Transformer. Using the DiT architecture, MDT~\cite{gao2023masked} introduced a masked latent modeling approach to effectively capture contextual information. In comparison to DiT, although MDT achieves faster learning speed and better FID scores on ImageNet-256 dataset~\cite{deng2009imagenet}, it has a more complex training pipeline. Recently with a similar architecture to DiT, SiT~\cite{ma2024sit} was proposed to incorporate flow matching~\cite{albergo2022building,albergo2023stochastic}. Unlike DiT, MDT or SiT, the proposed DiffiT does not use shift and scale, as in AdaLN formulation, for conditioning. Instead, DiffiT proposes a time-dependent self-attention (\textit{i.e.} TMSA) to jointly learn the spatial and temporal dependencies. In addition, DiffiT proposes both image and latent space models for different image generation tasks with different resolutions with SOTA performance.
\section{Methodology}
\label{sec:method}

\subsection{Diffusion Model Preliminaries}

Diffusion models~\cite{sohl-dickstein2015deep,ho2020denoising,song2020score} are a family of generative models that synthesize samples via an iterative denoising process. Given a data distribution as $q_0(\rvz_0)$, a family of random variables $\rvz_t$ for $t \in [0, T]$ are defined by injecting Gaussian noise to $\rvz_0$, \textit{i.e.}, $q_t(\rvz_t) = \int q(\rvz_t | \rvz_0) q_0(\rvz_0) \diff \rvz_0$, where $q(\rvz_t | \rvz_0) = \gN(\rvz_0, \sigma_t^2 \mI)$ is a Gaussian distribution. Typically, $\sigma_t$ is chosen as a non-decreasing sequence such that $\sigma_0 = 0$ and $\sigma_T$ being much larger than the data variance. This is called the ``Variance-Exploding'' noising schedule in the literature~\cite{song2020score}; for simplicity, we use these notations throughout the paper, but we note that it can be equivalently converted to other commonly used schedules (such as ``Variance-Preserving''~\cite{ho2020denoising}) by simply rescaling the data with a scaling term, dependent on $t$~\cite{song2020denoising,karras2022elucidating}. 

The distributions of these random variables are the marginal distributions of forward diffusion processes (Markovian or not~\cite{song2020denoising}) that gradually reduces the ``signal-to-noise'' ratio between the data and noise. As a generative model, diffusion models are trained to approximate the reverse diffusion process, that is, to transform from the initial noisy distribution (that is approximately Gaussian) to a distribution that is close to the data one.

\paragraph{Training} Despite being derived from different perspectives, diffusion models can generally be written as learning the following denoising autoencoder objective~\cite{vincent2011a}
\begin{align}
    \bb{E}_{q_0(\rvz_0), t \sim p(t), \epsilon \sim \gN(0, \mI)}[\lambda(t) \norm{\epsilon - \epsilon_\theta(\rvz_0 + \sigma_t \epsilon, t)}_2^2].
\end{align}
Intuitively, given a noisy sample from $q(\rvz_t)$ (generated via $\rvz_t := \rvz_0 + \sigma_t \epsilon$), a neural network $\epsilon_\theta$ is trained to predict the amount of noise added (\textit{i.e.}, $\epsilon$). Equivalently, the neural network can also be trained to predict $\rvz_0$ instead~\cite{ho2020denoising,salimans2022progressive}. The above objective is also known as denoising score matching~\cite{vincent2011a}, where the goal is to try to fit the data score (\textit{i.e.}, $\nabla_{\rvz_t} \log q(\rvz_t)$) with a neural network, also known as the score network $s_\theta(\rvz_t, t)$. The score network can be related to $\epsilon_\theta$ via the relationship $s_\theta(\rvz_t, t) := - \epsilon_\theta(\rvz_t, t) / \sigma_t$.
% (a known function of $\epsilon_\theta$). 

\paragraph{Sampling} Samples from the diffusion model can be simulated by the following family of stochastic differential equations that solve from $t = T$ to $t = 0$~\cite{grenander1994representations,karras2022elucidating,zhang2022gddim,dockhorn2021score}:
\begin{align}
    \diff \rvz = - (\dot{\sigma}_t + \beta_t) \sigma_t s_\theta(\rvz, t) \diff t + \sqrt{2 \beta_t} \sigma_t \diff \omega_t, \label{eq:score-sde}
\end{align}
where $\omega_t$ is the reverse standard Wiener process, and $\beta_t$ is a function that describes the amount of stochastic noise during the sampling process. If $\beta_t = 0$ for all $t$, then the process becomes a probabilistic ordinary differential equation~\cite{anderson1982reverse} (ODE), and can be solved by ODE integrators such as denoising diffusion implicit models (DDIM~\cite{song2020denoising}). Otherwise, solvers for stochastic differential equations (SDE) can be used, including the one for the original denoising diffusion probabilistic models (DDPM~\cite{ho2020denoising}). Typically, ODE solvers can converge to high-quality samples in fewer steps and SDE solvers are more robust to inaccurate score models~\cite{karras2022elucidating}. 
% \begin{figure}[t]
% \includegraphics[width=1.0\linewidth, clip=True, trim={2.7cm 6.0cm 12.0cm 6.8cm}]{Images/Diffit_block_bracket.pdf}
% \caption{The DiffiT Transformer block applies linear projection to spatial and time-embedding tokens before combining them together to form query, key, and value vectors for each token. These vectors are then used to compute multi-head self-attention activations, followed by two linear layers.}
% \label{fig:fig2}
% \end{figure}
% \begin{wrapfigure}{r}{0.5\textwidth}
%   \begin{center}
% \includegraphics[width=1.0\linewidth, clip=True, trim={2.7cm 6.0cm 12.0cm 6.8cm}]{Images/Diffit_block_bracket.pdf}

%   \end{center}
% \caption{The DiffiT Transformer block applies linear projection to spatial and time-embedding tokens before combining them together to form query, key, and value vectors for each token. These vectors are then used to compute multi-head self-attention activations, followed by two linear layers. Above, LN indicates Layer Norm~\cite{ba2016layer} and GELU denotes the Gaussian error linear unit activation function~\cite{hendrycks2016gaussian}. TMSA (time-dependent multi-head self-attention) and MLP (multi-layer perceptron) are discussed in Eq.~\ref{eq:tmsa} and Eq.~\ref{eq:mlp}. \vspace{-1.8cm}}
% \label{fig:fig2}
% \end{wrapfigure}

\subsection{DiffiT Model}
\label{sec:diffit_model}

\paragraph{Time-dependent Self-Attention} At every layer, our transformer block receives $\{\mathbf{x_s}\}$, a set of tokens arranged spatially on a 2D grid in its input. It also receives $\mathbf{x_t}$, a time token representing the time step. Similar to ~\cite{ho2020denoising}, 
we obtain the time token by feeding positional time embeddings to a small MLP with swish activation~\cite{elfwing2018sigmoid}. This time token is passed to all layers in our denoising network. We introduce our time-dependent multi-head self-attention, which captures both long-range spatial and temporal dependencies by projecting feature and time token embeddings in a shared space. %\js{for the sake of completeness, maybe add one line to talk about where these spatial and temporal features come from?} 
Specifically, time-dependent queries $\mathbf{q}$, keys $\mathbf{k}$ and values $\mathbf{v}$ in the shared space are computed by a linear projection of spatial and time embeddings $\mathbf{x_s}$ and $\mathbf{x_t}$ via
\begin{align}
\mathbf{q_s}    &= \mathbf{x_s} \mathbf{W}_{qs} + \mathbf{x_t} \mathbf{W}_{qt}, \label{eq:q} \\
\mathbf{k_s}    &= \mathbf{x_s} \mathbf{W}_{ks} + \mathbf{x_t} \mathbf{W}_{kt}, \label{eq:k} \\
\mathbf{v_s}    &= \mathbf{x_s} \mathbf{W}_{vs} + \mathbf{x_t} \mathbf{W}_{vt}, \label{eq:v}
\end{align}
where $\mathbf{W}_{qs}$, $\mathbf{W}_{qt}$, $\mathbf{W}_{ks}$, $\mathbf{W}_{kt}$, $\mathbf{W}_{vs}$, $\mathbf{W}_{vt}$ denote spatial and temporal linear projection weights for their corresponding queries, keys, and values respectively. 

We note that the operations listed in Eq.~\ref{eq:q} to \ref{eq:v} are equivalent to a linear projection of each spatial token, concatenated with the time token. As a result, key, query, and value are all linear functions of both time and spatial tokens and they can adaptively modify the behavior of attention for different time steps. We define $\mathbf{Q}:=\{\mathbf{q_s}\}$, $\mathbf{K}:=\{\mathbf{k_s}\}$, and $\mathbf{V}:=\{\mathbf{v_s}\}$ which are stacked form of query, key, and values in rows of a matrix. The self-attention is then computed as follows
\begin{align}
\attention(\mathbf{Q}, \mathbf{K}, \mathbf{V}) = \softmax\left(\frac{\mathbf{Q} \mathbf{K}^\top}{\sqrt{d}} + \mathbf{B} \right) \mathbf{V}. \label{eq:selfattn}
\end{align}
In which, $d$ is a scaling factor for keys $\mathbf{K}$, and $\textbf{B}$ corresponds to a relative position bias~\cite{shaw2018self}. For computing the attention, the relative position bias allows for the encoding of information across each attention head. Note that although the relative position bias is implicitly affected by the input time embedding, directly integrating it with this component may result in sub-optimal performance as it needs to capture both spatial and temporal information. Please see Sec.~\ref{sec:abl-temporal} for more analysis.

\begin{figure}[t]\centering
\includegraphics[width=1\linewidth]{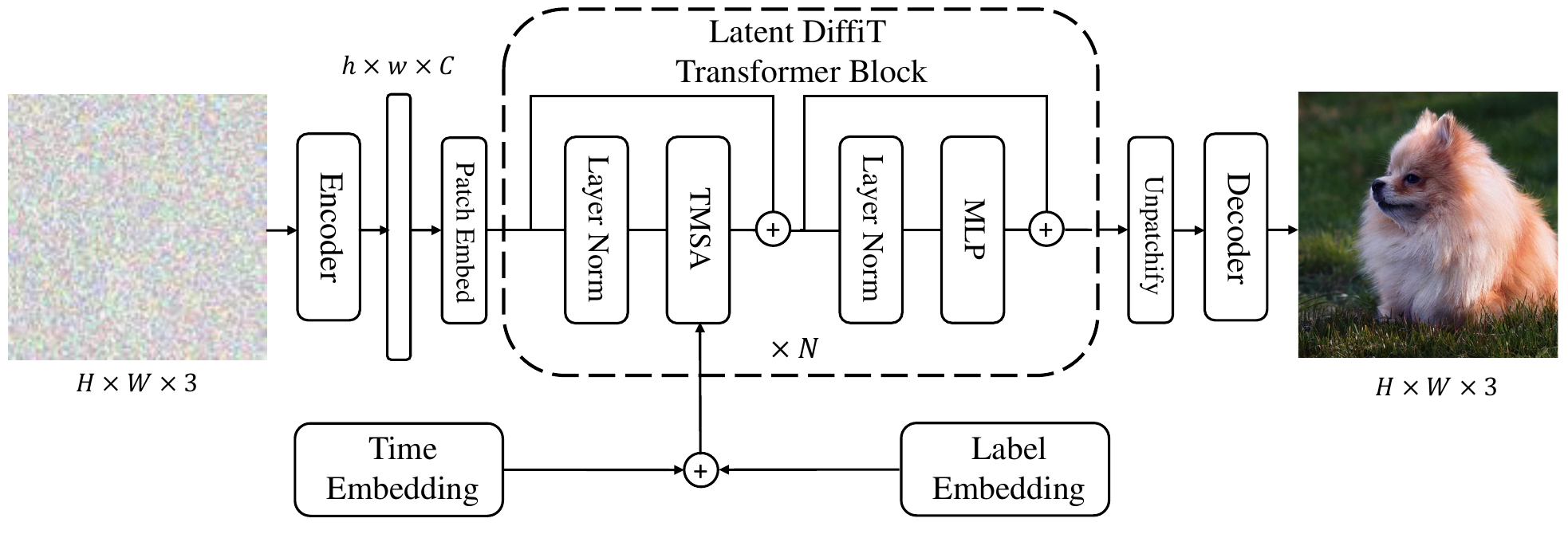}
\caption{Overview of the latent DiffiT framework.}\vspace{-4mm}
\label{fig:arch-framework-latent-diffit}
\end{figure}

\paragraph{DiffiT Transformer Block} The transformer block is a core building block of the proposed DiffiT architecture and is defined as
\iffalse
\begin{align}
    &{{\hat{\bf{x}}}^{l}} = \text{TMSA}\left( {\text{LN}\left( {{{\bf{x}}^{l - 1}}} \right)}, \bf{x_{t}} \right) + {\bf{x}}^{l - 1}, \label{eq:tmsa}\\
    &{{\bf{x}}^l} = \text{MLP}\left( {\text{LN}\left( {{{\hat{\bf{x}}}^{l}}} \right)} \right) + {{\hat{\bf{x}}}^{l}},\label{eq:mlp}
\end{align}
\fi
\begin{align}
    \bf{\hat{x}_s} &= \text{TMSA}\left( {\text{LN}\left( \bf{x_s} \right)}, \bf{x_{t}} \right) + \bf{x_s}, \label{eq:tmsa}\\
    \bf{x_s} &= \text{MLP}\left( {\text{LN}\left( \bf{\hat{x}_s} \right)} \right) + \bf{\hat{x}_s},\label{eq:mlp}
\end{align}
where TMSA denotes time-dependent multi-head self-attention, as described in the above, $\bf{x_{t}}$ is the time-embedding token, $\bf{x_{s}}$ is a spatial token, and LN and MLP denote Layer Norm~\cite{ba2016layer} and MLP respectively. 

\subsubsection{Latent Space}

Recently, latent diffusion models have been shown effective in generating high-quality large-resolution images~\cite{vahdat2021score, rombach2022high}. In Fig.~\ref{fig:arch-framework-latent-diffit}, we show the architecture of latent DiffiT model. We first encode the images using a pre-trained variational auto-encoder network~\cite{rombach2022high}. The feature maps are then converted into non-overlapping patches and projected into a new embedding space. Similar to the DiT model~\cite{peebles2022scalable}, we use a vision transformer, without upsampling or downsampling layers, as the denoising network in the latent space. In addition, we also utilize a three-channel classifier-free guidance to improve the quality of generated samples. The final stage is a linear layer to decode the output.

\subsubsection{Image Space}

\begin{figure*}[t]
    \centering
    \includegraphics[width=\textwidth]{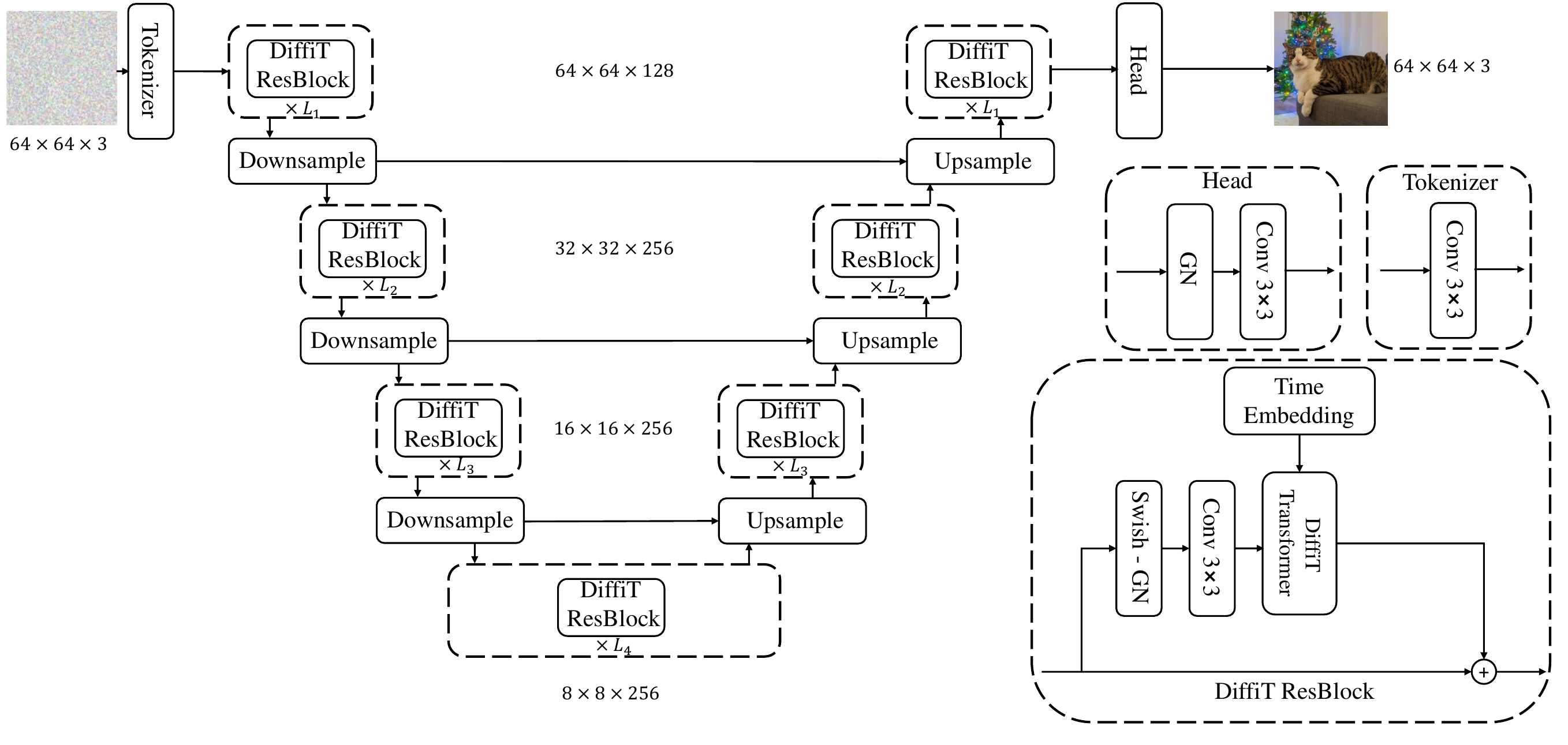}
    \caption{Overview of the image-space DiffiT model. Downsample and Upsample denote convolutional downsampling and upsampling layers, respectively. Please see the supplementary materials for more information regarding the DiffiT architecture.}
    \label{fig:arch-framework}
\end{figure*}

\paragraph{DiffiT Architecture} As shown in Fig.~\ref{fig:arch-framework}, DiffiT uses a symmetrical U-Shaped encoder-decoder architecture in which the contracting and expanding paths are connected to each other via skip connections at every resolution. Specifically, each resolution of the encoder or decoder paths consists of $L$ consecutive DiffiT blocks, containing our proposed time-dependent self-attention modules. In the beginning of each path, for both the encoder and decoder, a convolutional layer is employed to match the number of feature maps. A convolutional upsampling or downsampling layer is also used for transitioning between each resolution. We speculate that the use of these convolutional layers embeds inductive image bias that can further improve the performance. In the remainder of this section, we discuss the DiffiT Transformer block and our proposed time-dependent self-attention mechanism. We use our proposed Transformer block as the residual cells when constructing the U-shaped denoising architecture.

\paragraph{Local Attention} The quadratic cost of attention scales poorly when the number of spatial tokens is large, especially in the case of large feature maps. Without loss of generality, the above Transformer block can be applied to local regions, in which the self-attention is computed within non-overlapping partitioned windows. Although these partitioned windows do not allow information to be propagated between different regions, the U-Net structure with bottleneck layers permits information sharing between different regions.

\paragraph{DiffiT ResBlock} We define our final residual cell by combining our proposed DiffiT Transformer block with an additional convolutional layer in the form:
\begin{align}
    \bf{\hat{x}_s} &= \text{Conv}_{3\times3}\left(\text{Swish}\left(\text{GN}\left(\bf{x_s}\right) \right) \right), \label{eq:conv}\\
    \bf{x_s} &= \text{DiffiT-Block}\left( \bf{\hat{x}_s}, \bf{x_t} \right) + \bf{x_s},\label{eq:tran}
\end{align}
where GN denotes the group normalization operation~\cite{wu2018group} and DiffiT-Transformer is defined in Eq.~\ref{eq:tmsa} and Eq.~\ref{eq:mlp}. Our residual cell for image space diffusion models is a hybrid cell combining both a convolutional layer and our Transformer block.

% \begin{figure}[h]
%     \centering
%     \includegraphics[width=0.83\textwidth]{Images/latent_diffit.pdf}
%     \caption{Overview of the latent DiffiT framework.}
%     \label{fig:arch-framework-abl-v22}
% \end{figure}

% \begin{figure*}[t!]
% \centering

% \resizebox{0.8\linewidth}{!}{
% \begingroup
% \renewcommand*{\arraystretch}{0.3}
% \begin{tabular}{c}

% \includegraphics[clip=true, trim={0cm 0cm 0cm 128px}]{Images/cifar-32x32_this.png} \\
%     \vspace{1mm} \\

% \end{tabular}
% \endgroup
% }
% \caption{Visualization of uncurated generated images for CIFAR-10 dataset. Best viewed in color.}
% \label{fig:cifar}
% \end{figure*}

% Unlike DiT~\cite{peebles2022scalable}, our model does not use additional adaLN layers to incorporate the time dependency, as it leverages the proposed TMSA blocks for this purpose. Please see the supplementary materials for more details on the latent DiffiT architecture as well as the training setting.
\section{Results}
\label{sec:results}

\subsection{Latent Space}
\begin{figure*}[htb]
\centering\includegraphics[width=\textwidth]{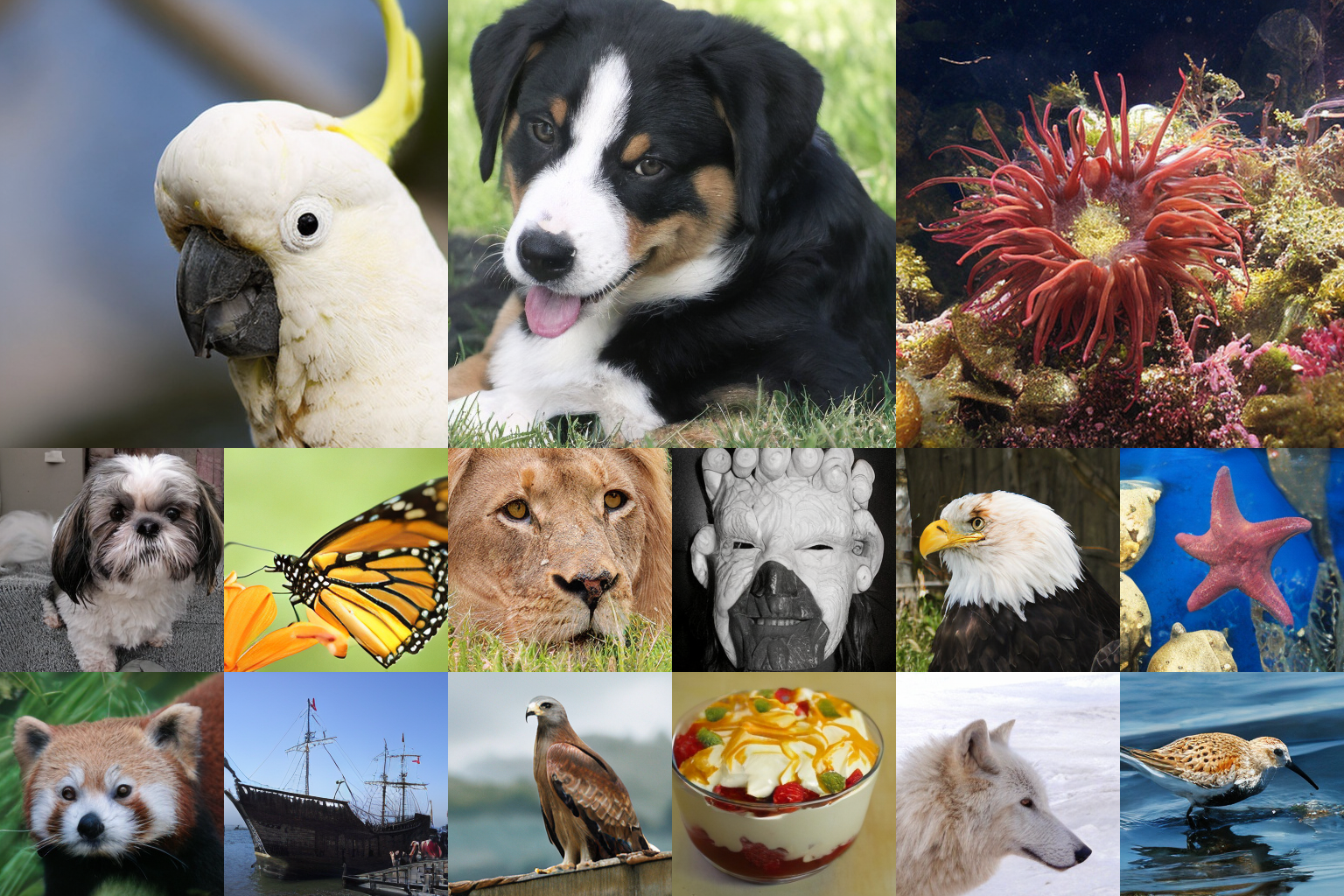}
\caption{Visualization of uncurated generated images on ImageNet-256 and ImageNet-512 datasets by latent DiffiT model.}\label{fig:imagenet-256
}
\label{fig:imagenet-256}
\end{figure*}

We have trained the latent DiffiT model on ImageNet-512 and ImageNet-256 dataset respectively. In Table.~\ref{tab:imagenet}, we present a comparison against other approaches using various image quality metrics. For this comparison, we select the best performance metrics from each model which may include techniques such as classifier-free guidance. In ImageNet-256 dataset, the latent DiffiT model outperforms competing approaches, such as SiT-XL~\cite{ma2024sit}, MDT-G~\cite{gao2023masked}, DiT-XL/2-G~\cite{peebles2022scalable} and StyleGAN-XL~\cite{sauer2022stylegan}, in terms of FID score and sets a new SOTA FID score of \textbf{1.73}. In terms of other metrics such as IS and sFID, the latent DiffiT model shows a competitive performance, hence indicating the effectiveness of the proposed time-dependent self-attention. In the ImageNet-512 dataset, the latent DiffiT model significantly outperforms DiT-XL/2-G in terms of both FID and Inception Score (IS). Although StyleGAN-XL~\cite{sauer2022stylegan} shows better performance in terms of FID and IS, GAN-based models are known to suffer from issues such as low diversity that are not captured by the FID score. These issues are reflected in sub-optimal performance of StyleGAN-XL in terms of both Precision and Recall. In addition, in Fig.~\ref{fig:imagenet-256}, we show a visualization of uncurated images that are generated on ImageNet-256 and ImageNet-512 dataset. We observe that the latent DiffiT model is capable of generating diverse high quality images across different classes.   

\begin{table*}[t]
\centering
\resizebox{.95\linewidth}{!}{
\begin{tabular}{lcccccccccc}
\toprule
\multirow{2}{*}{\textbf{Model}} &
\multirow{2}{*}{\textbf{Class}} &
\multicolumn{4}{c}{\textbf{ImageNet-256}} & \multicolumn{5}{c}{\textbf{ImageNet-512}}\\
\cmidrule{3-6}\cmidrule{8-11}
&& FID $\downarrow$ & IS $\uparrow$  & Precision $\uparrow$ & Recall $\uparrow$&  & FID $\downarrow$ & IS $\uparrow$  & Precision $\uparrow$ & Recall $\uparrow$\\
\midrule
LDM-4~\cite{rombach2022high} & Diffusion &10.56 & 103.49 & 0.71&\textbf{0.62} & &- & - &  - &  - \\
BigGAN-Deep~\cite{brock2018large} & GAN &6.95 & 171.40 & \textbf{0.87}& 0.28 & &8.43 & 177.90 &  \textbf{0.88} &  0.29 \\
MaskGIT~\cite{chang2022maskgit} & Masked Modeling &4.02 & \textbf{355.60} & \underline{0.83} & 0.44 & &4.46 & \textbf{342.00} &  0.83 &  0.50 \\
RQ-Transformer~\cite{lee2022autoregressive} & Autoregressive &3.80 & \underline{323.70} & - & - & &- & - &  - &  - \\
ADM-G-U~\cite{dhariwal2021diffusion} & Diffusion &3.94 & 215.84 & \underline{0.83} & 0.53 & &3.85 & 221.72 &  \underline{0.84} &  \underline{0.53} \\
LDM-4-G~\cite{rombach2022high} & Diffusion &3.60 & 247.67 & \textbf{0.87} & 0.48 & &- & - &  - &  -  \\
Simple Diffusion~\cite{hoogeboom2023simple} & Diffusion &2.77 & 211.80 & - & - & &3.54 & 205.30 &  - &  -  \\
DiT-XL/2-G~\cite{peebles2022scalable} & Diffusion &2.27 & \underline{278.24} & \underline{0.83} & 0.57 & &3.04 & 240.82 &  \underline{0.84} &  0.54  \\
StyleGAN-XL~\cite{sauer2022stylegan} & GAN &2.30 & 265.12 & 0.78 & 0.53 & &\textbf{2.41} & \underline{267.75} &  0.77 & 0.52  \\
MDT-G~\cite{gao2023masked} & Diffusion &\underline{1.79} & 283.01 & 0.81 & \underline{0.61} & &- & - &  - &  -  \\
SiT-XL~\cite{ma2024sit} & Diffusion &2.06 & 270.27 & 0.82 & 0.59 & &- & - &  - &  -  \\
\hline
\rowcolor{Gray}
\textbf{DiffiT} & Diffusion & \textbf{1.73} & 276.49 & 0.80 & \textbf{0.62} & &\underline{2.67} & 252.12 &  0.83 & \textbf{0.55}  \\

\bottomrule
\end{tabular}}
\caption{Comparison of image generation performance against state-of-the-art models on ImageNet-256 and ImageNet-512 dataset. The latent DiffiT model achieves SOTA performance in terms of FID score on ImageNet-256 dataset.
}
\label{tab:imagenet}
\end{table*}

\subsection{Image Space}
\begin{wraptable}{r}{6cm}
\vspace{-12.2mm}
\centering
\setlength{\tabcolsep}{1pt}
\caption{FID performance comparison against various generative approaches on the CIFAR10, FFHQ-64 datasets. VP and VE denote Variance Preserving and Variance Exploding respectively.}\label{tab:cifar10_main}
\resizebox{1\linewidth}{!}{
\begin{tabular}{lcccc}
        \toprule
        {\bf Method} & {\bf Class}& {\bf Space Type}   & {\bf CIFAR-10} & {\bf FFHQ}   \\
                     & && 32$\times$32 & 64$\times$64    \\
        \midrule
        NVAE \cite{vahdat2020nvae}      & VAE & - & 23.50 & -    \\
        GenViT~\cite{yang2022your}      & Diffusion & Image & 20.20 & -   \\
        AutoGAN \cite{gong2019autogan}      & GAN & - & 12.40 & -   \\
        TransGAN \cite{jiang2021transgan}      & GAN & - & 9.26 & -    \\
        INDM~\cite{kim2022maximum}      & Diffusion & Latent & 3.09 & -   \\
        DDPM++ (VE)~\cite{song2020score}      & Diffusion & Image & 3.77 & 25.95    \\
        U-ViT~\cite{bao2022all}      & Diffusion & Image & 3.11 & -  \\
        DDPM++ (VP)~\cite{song2020score}      & Diffusion & Image & 3.01 & 3.39    \\
        StyleGAN2 w/ ADA \cite{karras2020training}      & GAN & - & 2.92 & -   \\
        LSGM~\cite{vahdat2021score}      & Diffusion & Latent & 2.01 & -   \\
        EDM (VE)~\cite{karras2022elucidating}      & Diffusion & Image & 2.01 & 2.53    \\
        EDM (VP)~\cite{karras2022elucidating}      & Diffusion & Image & 1.99 & 2.39   \\
        \hline
        \rowcolor{Gray}
        \textbf{DiffiT} (Ours)      & Diffusion & Image & \textbf{1.95} & \textbf{2.22}    \\
        \midrule
    \end{tabular}%
    }
    \vspace{-1mm}
\end{wraptable}
We have trained the image space DiffiT model on FFHQ-64~\cite{karras2019style} and CIFAR10~\cite{krizhevsky2009learning} datasets. In Table.~\ref{tab:cifar10_main}, we compare the performance of our model against a variety of different generative models including other score-based diffusion models as well as GANs, and VAEs. DiffiT achieves a state-of-the-art image generation FID score of 1.95 on the CIFAR-10 dataset, outperforming state-of-the-art diffusion models such as EDM~\cite{karras2022elucidating} and LSGM~\cite{vahdat2021score}. In comparison to two recent ViT-based diffusion models, our proposed DiffiT significantly outperforms U-ViT~\cite{bao2022all} and GenViT~\cite{yang2022your} models in terms of FID score in CIFAR-10 dataset. Additionally, DiffiT significantly outperforms EDM~\cite{karras2022elucidating} and DDPM++~\cite{song2020score} models, both on VP and VE training configurations, in terms of FID score. In Fig.~\ref{fig:ffhq}, we illustrate the generated images on FFHQ-64 dataset. Please see supplementary materials for CIFAR-10 generated images.

% \twocolumn[{
% \renewcommand\twocolumn[1][]{#1}%
% \begin{center}
%     \includegraphics[width=\textwidth]{Images/imagenet/imagenet_comb.png}
%     \captionof{figure}{Visualization of uncurated generated 256$\times$256 images on ImageNet-256 and ImageNet-512 datasets}\label{fig:imagenet-256}
% \end{center}
% }]

\begin{figure*}[t]
\centering

\resizebox{\linewidth}{!}{
\begingroup
\renewcommand*{\arraystretch}{0.3}
\begin{tabular}{c}

\includegraphics[clip=True, trim={0cm 64px, 0cm 384px}]{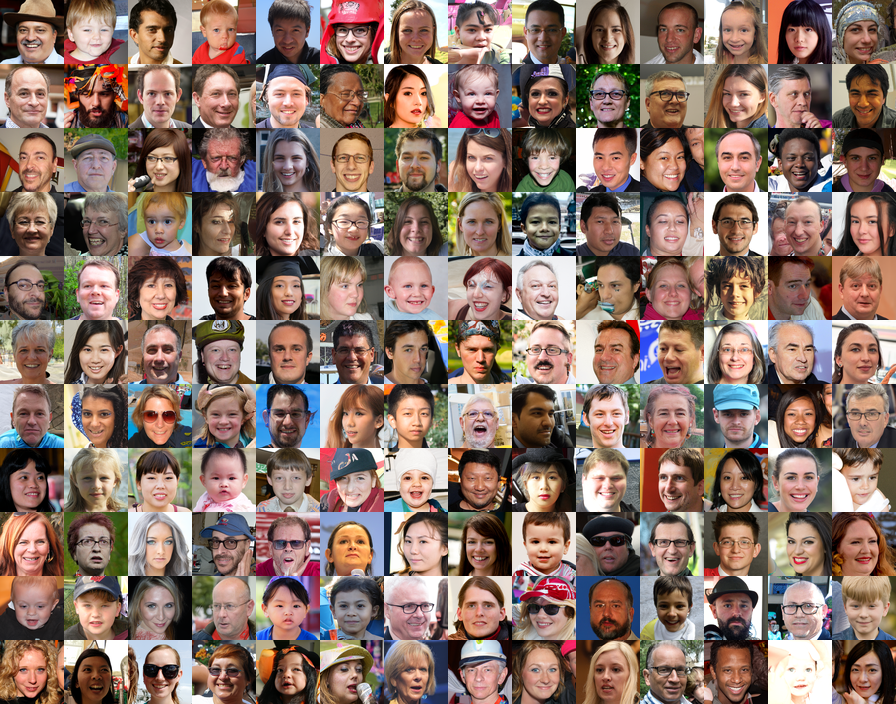} \\
    \vspace{1mm} \\

\end{tabular}
\endgroup
}

\caption{Visualization of uncurated generated images for FFHQ-64 dataset. Best viewed in color.}
\label{fig:ffhq}
\end{figure*}

\section{Ablation}
\label{sec:ablation}
In this section, we provide additional ablation studies to provide insights into DiffiT. We address different questions such as: (1) What strikes the right balance between time and feature token dimensions ? (2) How do different components of DiffiT contribute to the final generation performance, (3) What is the optimal way of introducing time dependency in our Transformer block? and (4) How does our time-dependent attention behave as a function of time?

\subsection{Time and Feature Token Dimensions}
\label{sec:abl-effect-time}
\begin{wraptable}{r}{7.5cm}
\vspace{-7.7mm}
\centering
\resizebox{.72\linewidth}{!}{
\setlength{\tabcolsep}{2.5pt}
  \begin{tabular}{lccc}
    \toprule
    Time Dimension  & Dimension   & CIFAR10 & FFHQ64 \\
    \midrule	 
    512&512  & 1.99 &2.27 \\
    256&256  & 2.13 &2.41 \\\rowcolor{Gray}
    512&512  & 1.95 &2.22 \\
    \bottomrule
    
  \end{tabular} 
  }
    \caption{Ablation study on the effectiveness of time and feature dimensions.}
    \label{tab:abl-1-time}
    \vspace{-7mm}
\end{wraptable}We conduct experiments to study the effect of the size of time and feature token dimensions on the overall performance. As shown below, we observe degradation of performance when the token dimension is increased from 256 to 512. Furthermore, decreasing the time embedding dimension from 512 to 256 impacts the performance negatively. 

\subsection{Effect of Architecture Design}
As presented in Table~\ref{tab:abl-1}, we study the effect of various components of both encoder and decoder in the architecture design on the image generation performance in terms of FID score on CIFAR-10. For these experiments, the projected temporal component is adaptively scaled and simply added to the spatial component in each stage. We start from the original ViT~\cite{dosovitskiy2020image} base model with 12 layers and employ it as the encoder (config A). For the decoder, we use the Multi-Level Feature Aggregation variant of SETR~\cite{zheng2021rethinking} (SETR-MLA) to generate images in the input resolution. Our experiments show this architecture is sub-optimal as it yields a final FID score of 5.34. We hypothesize this could be due to the isotropic architecture of ViT which does not allow learning representations at multiple scales. We then extend the encoder ViT into 4 different\label{sec:abl-effect}
\begin{wraptable}{r}{7.5cm}
\vspace{-7mm}
\centering
\resizebox{.88\linewidth}{!}{
\setlength{\tabcolsep}{2.5pt}
  \begin{tabular}{lccc}
    \toprule
    Config & Encoder   & Decoder & FID Score \\
    \midrule	 
    A&ViT~\cite{dosovitskiy2020image}  & SETR-MLA~\cite{zheng2021rethinking} &5.34 \\
   B& + Multi-Resolution   & SETR-MLA~\cite{zheng2021rethinking} &4.64 \\
   C& Multi-Resolution   & + Multi-Resolution  &3.71 \\
    D&+ DiffiT Encoder & Multi-Resolution   & 2.27 \\ \rowcolor{Gray}
    E&+ DiffiT Encoder & + DiffiT Decoder & \textbf{1.95}\\
    \bottomrule
    
  \end{tabular} 
  }
    \caption{Ablation study on the effectiveness of encoder and decoder architecture.}
    \label{tab:abl-1}
    \vspace{-7mm}
\end{wraptable} multi-resolution stages with a convolutional layer in between each stage for downsampling (config B). We denote this setup as Multi-Resolution and observe that these changes and learning multi-scale feature representations in the encoder substantially improve the FID score to 4.64. In addition, instead of SETR-MLA~\cite{zheng2021rethinking} decoder, we construct a symmetric U-like architecture by using the same Multi-Resolution setup except for using convolutional layers between stages for upsampling (config C). These changes further improve the FID score to 3.71. Furthermore, we first add the DiffiT Transformer blocks and construct a \begin{wraptable}{r}{7.5cm}
\vspace{-7.3mm}
\centering
\resizebox{.67\linewidth}{!}{
\setlength{\tabcolsep}{2.5pt}
  \begin{tabular}{lcc}
    \toprule
    Model & TMSA   & FID Score \\
    \midrule	 
    DDPM++(VE)~\cite{song2020score}&No  & 3.77 \\\rowcolor{Gray}
    DDPM++(VE)~\cite{song2020score}&Yes  & 3.49 \\
    DDPM++(VP)~\cite{song2020score}&No  & 3.01 \\\rowcolor{Gray}
    DDPM++(VP)~\cite{song2020score}&Yes  & 2.76 \\

    \bottomrule
  \end{tabular} 
  }
    \caption{Effectiveness of TMSA.}
    \label{tab:attn_impact}
    \vspace{-7mm}
\end{wraptable}DiffiT Encoder and observe that FID scores substantially improve to 2.27 (config D). As a result, this validates the effectiveness of the proposed TMSA in which the self-attention models both spatial and temporal dependencies. Using the DiffiT decoder further improves the FID score to 1.95 (config E), hence demonstrating the importance of DiffiT Transformer blocks for decoding.

\subsection{Time-Dependent Self-Attention}
\label{sec:abl-temporal-effect}
We evaluate the effectiveness of our proposed TMSA layers in a generic denoising network. Specifically, using the DDPM++~\cite{song2020score} model, we replace the original self-attention layers with TMSA layers for both VE and VP settings for image generation on the CIFAR10 dataset. Note that we did not change the 
original hyper-parameters for this study. As shown in Table~\ref{tab:attn_impact} employing TMSA decreases the FID scores by 0.28 and 0.25 for VE and VP settings respectively. These results demonstrate the effectiveness of the proposed TMSA to dynamically adapt to different sampling steps and capture temporal information.   

\subsection{Impact of Self-Attention Components}
\label{sec:abl-temporal}
\begin{wraptable}{r}{7.5cm}
\vspace{-7.7mm}
\centering
\resizebox{.67\linewidth}{!}{
\setlength{\tabcolsep}{2.5pt}
  \begin{tabular}{lcc}
    \toprule
    Config & Component   & FID Score \\
    \midrule	 
    F&Relative Position Bias  & 3.97 \\
    G&MLP  & 3.81 \\\rowcolor{Gray}
   H& TMSA  & \textbf{1.95} \\

    \bottomrule
    
  \end{tabular} 
  }
    \caption{Effectiveness of different components.}
    \label{tab:abl-2}
    \vspace{-3mm}
\end{wraptable}
In Table~\ref{tab:abl-2}, we study different design choices for introducing time-dependency in self-attention layers. In the first baseline, we remove the temporal component from our proposed TMSA and we only add the temporal tokens to relative positional bias (config F). We observe a significant increase in the FID score to 3.97 from 1.95. In the second baseline, instead of using relative positional bias, we add temporal tokens to the MLP layer of DiffiT Transformer block (config G). We observe that the FID score slightly improves to 3.81, but it is still suboptimal compared to our proposed TMSA (config H). Hence, this experiment validates the effectiveness of our proposed TMSA that integrates time tokens directly with spatial tokens when forming queries, keys, and values in self-attention layers.

\subsection{Time Token in TMSA}
\label{sec:abl-time-embed-tmsa}
\begin{wraptable}{r}{7.5cm}
\vspace{-7.6mm}
\centering
\resizebox{.67\linewidth}{!}{
\setlength{\tabcolsep}{2.5pt}
  \begin{tabular}{lccc}
    \toprule
    	Model  & TMSA Design   & CIFAR10 & FFHQ64 \\
    \midrule	 
    DiffiT&Separate  & 2.28 &2.59 \\\rowcolor{Gray}
    DiffiT&Mixed  & \textbf{1.95} &\textbf{2.22} \\
    \bottomrule
    
  \end{tabular} 
  }
    \caption{Impact of time embedding.}
    \label{tab:abl-2-time-tmsa}
    \vspace{-7mm}
\end{wraptable}
We investigate if treating time embedding as a separate token in TMSA is a beneficial choice. Specifically, we apply self-attention to spatial and time tokens separately to understand the impact of decoupling them. As shown in Table~\ref{tab:abl-2-time-tmsa}, we observe the degradation of performance for CIFAR10, FFHQ64 datasets, in terms of FID score. Hence, the decoupling of spatial and temporal information in TMSA leads to suboptimal performance.

\subsection{Time Embedding}
\label{sec:abl-time-embed}
\begin{wraptable}{r}{7.5cm}
\vspace{-7.4mm}
\centering
\resizebox{.67\linewidth}{!}{
\setlength{\tabcolsep}{2.5pt}
  \begin{tabular}{lccc}
    \toprule
    	Model  & Time Embedding   & CIFAR10 & FFHQ64 \\
    \midrule	 
    DiffiT&Fourier  & 2.02 &2.37 \\\rowcolor{Gray}
    DiffiT&Positional  & \textbf{1.95} &\textbf{2.22} \\
    \bottomrule
    
  \end{tabular} 
  }
    \caption{Effectiveness of TMSA time token.}
    \label{tab:abl-2-time}
    \vspace{-7mm}
\end{wraptable}We study the sensitivity of the DiffiT model to different time embeddings such as Fourier and positional time embeddings. As shown in Table~\ref{tab:abl-2-time}, using a Fourier time embedding leads to degradation of performance in terms of FID score for both CIFAR10~\cite{krizhevsky2009learning} and FFHQ-64~\cite{karras2019style} datasets.

\begin{figure}[t]
  \centering
  \begin{subfigure}{0.49\linewidth}
   \includegraphics[width=0.98\linewidth]{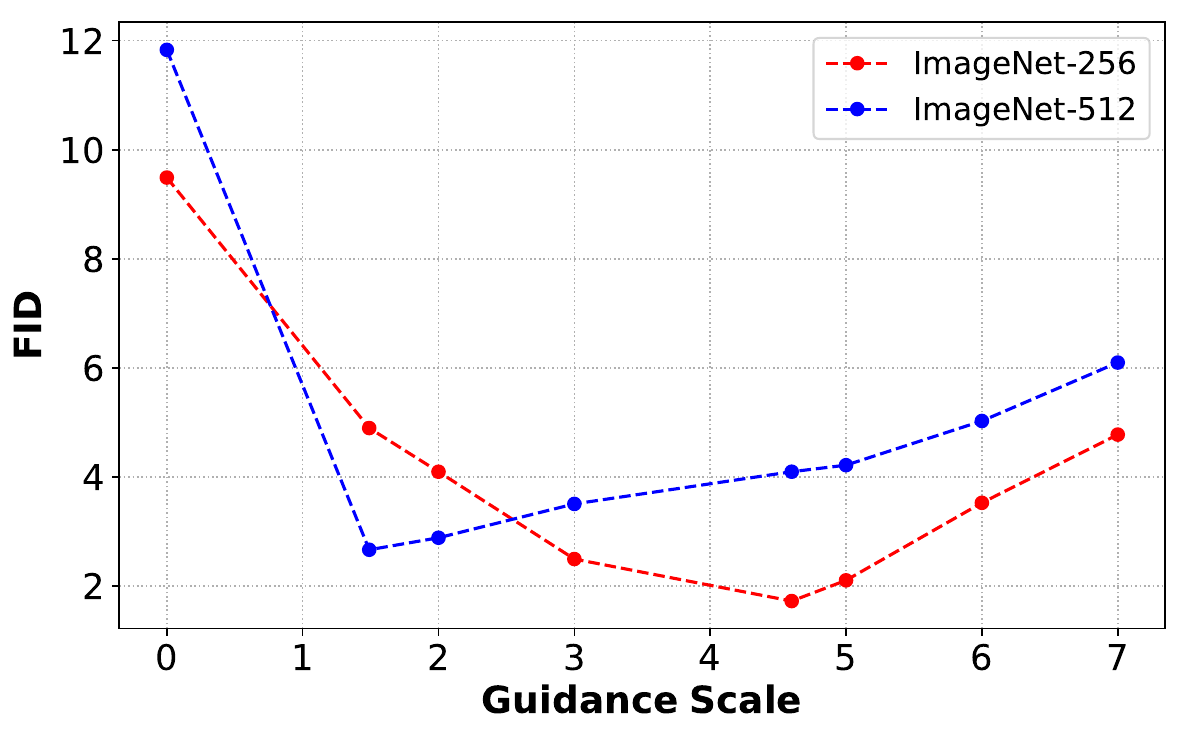}
    \caption{Effect of classifier-free guidance.}
    \label{fig:short-a}
  \end{subfigure}
  \hfill
  \begin{subfigure}{0.49\linewidth}
   \includegraphics[width=0.98\linewidth]{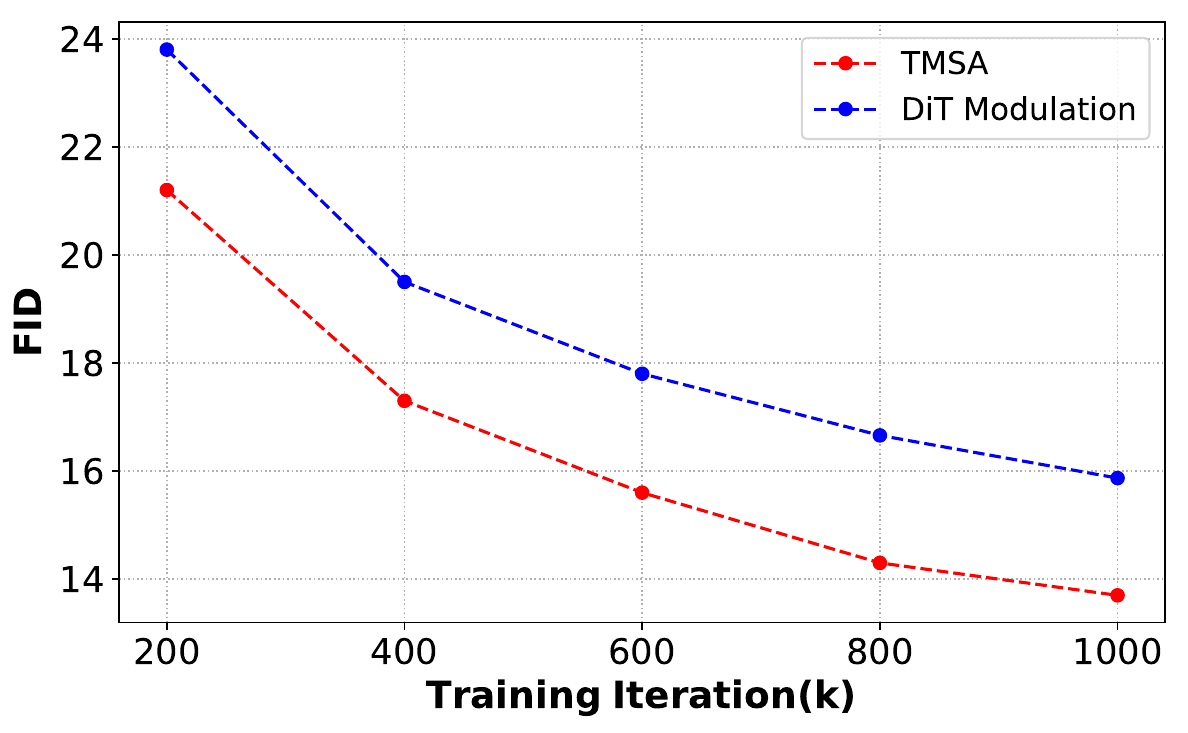}
    \caption{TMSA vs DiT modulation comparison.}
    \label{fig:short-b}
  \end{subfigure}
  \caption{(a) Effect of classifier-free guidance scale on FID score for ImageNet-256 and ImageNet-512. (b) Performance comparison of TMSA and DiT modulation.}
  \label{fig:guideance}
\end{figure}

% DiT-XL/2-G~\cite{peebles2022scalable}
\begin{wraptable}{r}{7.5cm}
\vspace{-7.7mm}
\centering
\resizebox{.67\linewidth}{!}{
\setlength{\tabcolsep}{2.5pt}
    \begin{tabular}{lccc}
    \toprule
        Model    & Parameters (M) & FLOPs (G) & FID \\
    \midrule
    DiT-XL/2-G~\cite{peebles2022scalable}    & 675  & 119  & 2.27  \\
    SiT-XL~\cite{ma2024sit}     & 675  & 119  & 2.06  \\
    MDT-G~\cite{gao2023masked}     & 700  & 121  & 1.79  \\
    \rowcolor{Gray}
    \bf DiffiT     & \bf 561  & \bf 114  & \bf 1.73  \\
    \bottomrule
    \end{tabular}
  }
    \caption{Computational efficiency comparison.}
    \label{tab:model_param_count}
    \vspace{-7mm}
\end{wraptable}
\subsection{Computational Efficiency}
\label{sec:abl-model-param}
In Table~\ref{tab:model_param_count}, we study the significance of model capacity in generating high-quality images by comparing the number of parameters for models that are trained on ImageNet-256 dataset. All models use the same number of function evaluations for sample generation for fair comparisons. We also use the same global window size for computing self-attention. We observe that DiffiT has $19.85\%$, $16.88\%$ and $16.88\%$ less number of parameters and $6.14\%$, $4.38\%$ and $4.38\%$ less number of FLOPs in comparison to MDT-G, SiT-XL and DiT-XL/2-G models, respectively while demonstrating a better FID score.

\subsection{Effect of Classifier-Free Guidance}
\label{sec:abl-class-guidance}
As shown in Fig.~\ref{fig:guideance} (a), we investigate the effect of classifier-free guidance scale on the quality of generated samples in terms of FID score. For the ImageNet-256 experiment, we used the improved classifier-free guidance~\cite{gao2023masked} which uses a power-cosine schedule to increase the diversity of generated images in early sampling stages. This scheme was not used for the ImageNet-512 experiment, since it did not result in any significant improvements. The guidance scales of 4.6 and 1.49 correspond to best FID scores of 1.73 and 2.67 for ImageNet-256 and ImageNet-512 experiments, respectively. Increasing the guidance scale beyond these values results in degradation of FID score.

\subsection{TMSA and DiT Modulation}
\label{sec:abl-dit}
We directly compare the performance of TMSA and DiT modulation mechanisms in Fig.~\ref{fig:guideance} (b). For this purpose, we employ a DiT-XL as the base model with TMSA as well as its original modulation and train both models for 1000K iterations on ImageNet-256 dataset. The model with TMSA consistently shows better FID scores in different training iterations, hence validating the effectiveness of TMSA.

\subsection{Effect of Window Size}
\label{sec:abl-window-size}
\begin{wrapfigure}{r}{0.58\textwidth}
\vspace{-18.5mm}
  \begin{center}
\includegraphics[width=0.85\linewidth]{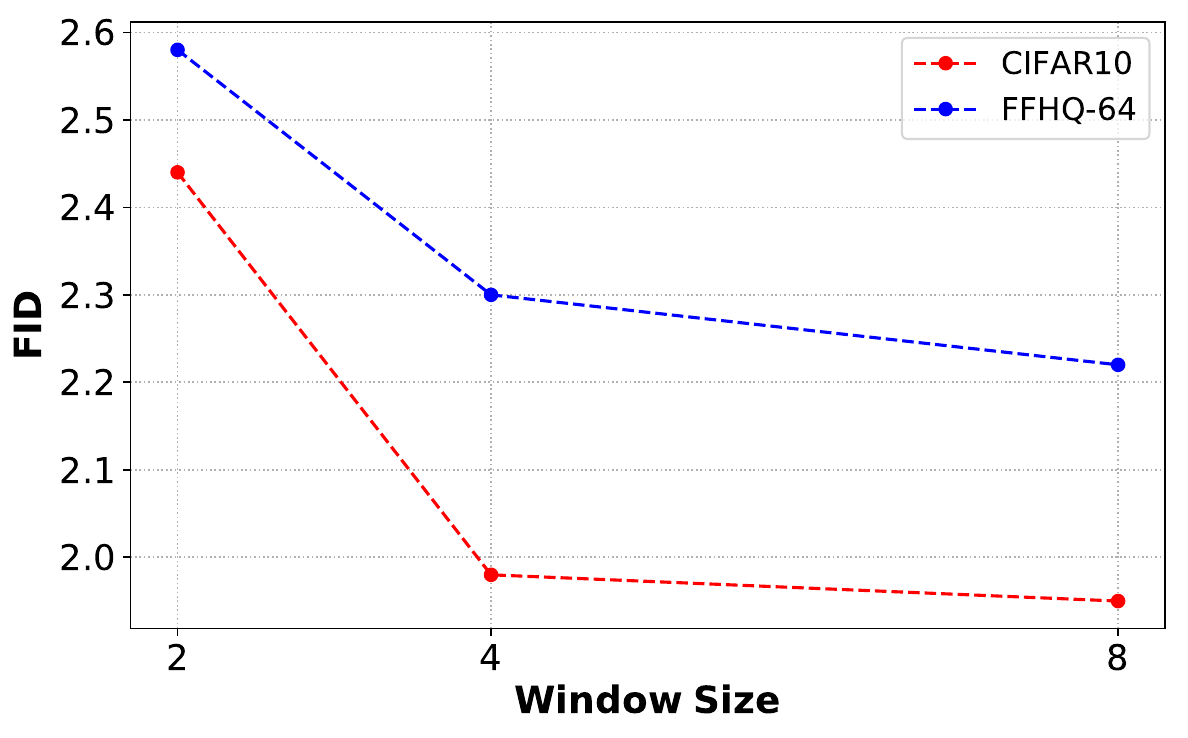}
  \end{center}
\caption{Impact of TMSA window size on FID.}\label{fig:window_size_eff}
\vspace{-2.5mm}
\end{wrapfigure}
As illustrated in Fig.~\ref{fig:window_size_eff}, we study the impact of window size in TMSA on the FID score of generated images for models that are trained on CIFAR10 ($32 \times 32$ resolution) and FFHQ-64 ($64 \times 64$ resolution) datasets. Increasing the TMSA window size from $2$ to $4$ decreases the FID score by $23.23\%$ and $12.17\%$ for CIFAR10 and FFHQ-64 models, respectively. As expected, increasing the effective receptive field seems to improve the generation performance. However, increasing the window size further from $4$ to $8$ only results in marginal improvement of $1.53\%$ and $3.60\%$ for CIFAR10 and FFHQ-64, respectively. This is due to the spatial redundancy of adjacent pixels which may not contribute significantly to the generation quality upon increasing the receptive field. As also discussed in the supplementary materials, for image space experiments, we have used our window-based TMSA formulation to benefit from the efficiency gains while maintaining high image generation quality.

 \section{Conclusion}
\label{sec:conclusion}

In this work, we presented DiffiT which is a novel ViT-based diffusion model for both latent and image space generation tasks. Specifically, we proposed the TMSA which allows self-attention to dynamically adapt to different stages of denoising while learning spatial and temporal dependencies and their interaction. The proposed TMSA also significantly improves the parameter efficiency. DiffiT achieves a new SOTA performance on ImageNet-256 dataset while having significantly less number of parameters in comparison to other competitive Transformer-based diffusion models such as SiT, MDT and DiT.

\bibliographystyle{splncs04}
\bibliography{main}

\newpage
\section*{Appendix}
\renewcommand{\thesection}{\Alph{section}}
\renewcommand\thefigure{S.\arabic{figure}}
\setcounter{figure}{0}
\renewcommand\thetable{S.\arabic{table}}
\setcounter{table}{0}

\section{Ablation}
\label{sec:abl_supp}

\subsection{Comparison to DiT and LDM}
On contrary to LDM~\cite{rombach2022high} and DiT~\cite{peebles2022scalable}, the latent DiffiT does not rely on shift and scale, as in AdaLN~\cite{peebles2022scalable}, or concatenation to incorporate time embedding into the denoising networks. However, DiffiT uses a time-dependent self-attention (\textit{i.e.} TMSA) to jointly learn the spatial and temporal dependencies. In addition, DiffiT proposes both image and latent space models for different image generation tasks with different resolutions with SOTA performance. Specifically, as shown in Table~\ref{tab:imagenet-supp}, DiffiT significantly outperforms LDM~\cite{rombach2022high} and DiT~\cite{peebles2022scalable} by 31.26\% and 51.94\% in terms of FID score on ImageNet-256~\cite{deng2009imagenet} dataset. In addition, DiffiT outperforms DiT~\cite{peebles2022scalable} by 13.85\% on ImageNet-512~\cite{deng2009imagenet} dataset. Hence, these benchmarks validate the effectiveness of the proposes architecture and TMSA design in DiffiT model as opposed to previous SOTA for both CNN and Transformer-based diffusion models.    

\begin{table*}[h]
\centering
\resizebox{.95\linewidth}{!}{
\begin{tabular}{lcccccccccc}
\toprule
\multirow{2}{*}{\textbf{Model}} &
\multirow{2}{*}{\textbf{Class}} &
\multicolumn{4}{c}{\textbf{ImageNet-256}} & \multicolumn{5}{c}{\textbf{ImageNet-512}}\\
\cmidrule{3-6}\cmidrule{8-11}
&& FID $\downarrow$ & IS $\uparrow$  & Precision $\uparrow$ & Recall $\uparrow$&  & FID $\downarrow$ & IS $\uparrow$  & Precision $\uparrow$ & Recall $\uparrow$\\
\midrule
LDM-4-G~\cite{rombach2022high} & Diffusion &3.60 & 247.67 & 0.87 & 0.48 & &- & - &  - &  -  \\
DiT-XL/2-G~\cite{peebles2022scalable} & Diffusion &2.27 & 278.24 & 0.83 & 0.57 & &3.04 & 240.82 &  0.84 &  0.54  \\
\rowcolor{Gray}
\textbf{DiffiT} & Diffusion & \textbf{1.73} & 276.49 & 0.80 & 0.62 & &\textbf{2.67} & 252.12 &  0.83 & 0.55  \\

\bottomrule
\end{tabular}}
\caption{Comparison of image generation performance against state-of-the-art models on ImageNet-256 and ImageNet-512 dataset. The latent DiffiT model achieves SOTA performance in terms of FID score on ImageNet-256 dataset.
}
\label{tab:imagenet-supp}
\end{table*}

\section{Architecture}
\label{sec:arch}
\subsection{Image Space}
We provide the details of blocks and their corresponding output sizes for both the encoder and decoder of the DiffiT model in Table~\ref{tab:encoder} and Table~\ref{tab:decoder}, respectively. The presented architecture details denote models that are trained with 64$\times$64 resolution. Without loss of generality, the architecture can be extended for 32$\times$32 resolution. For FFHQ-64~\cite{karras2019style} dataset, the values of $L_{1}$, $L_{2}$, $L_{3}$ and $L_{4 }$ are 4, 4, 4, and 4 respectively. For CIFAR-10~\cite{krizhevsky2009learning} dataset, the architecture spans across three different resolution levels (\textit{i.e.} 32, 16, 8), and the values of $L_{1}$, $L_{2}$, $L_{3}$ are 4, 4, 4 respectively. Please refer to the paper for more information regarding the architecture details.

\begin{table}[t] \centering
\setlength{\tabcolsep}{4pt}
\caption{Detailed description of components in DiffiT encoder for models that are trained at $64\times64$ resolution.}
\label{tab:encoder}
\begin{tabular}{ll}
\toprule
Component Description & Output size    \\ 
\midrule
Input & $64\times64\times3$   \\
Tokenizer & $64\times64\times128$    \\
DiffiT ResBlock $\times L_{1}$ & $64\times64\times128$    \\
Downsampler & $32\times32\times128$    \\
DiffiT ResBlock $\times L_{2}$ & $32\times32\times256$    \\
Downsampler & $16\times16\times128$    \\
DiffiT ResBlock $\times L_{3}$ & $16\times16\times256$    \\
Downsampler & $8\times8\times256$    \\
DiffiT ResBlock $\times L_{4}$ & $8\times8\times256$    \\
\bottomrule
\end{tabular}
\end{table}

\begin{table}[t] \centering
\setlength{\tabcolsep}{4pt}
\caption{Detailed description of components in DiffiT decoder for models that are trained at $64\times64$ resolution.}
\label{tab:decoder}
\begin{tabular}{ll}
\toprule
Component Description & Output size    \\ 
\midrule
Input & $8\times8\times256$   \\
Upsampler & $16\times16\times256$    \\
DiffiT ResBlock $\times L_{3}$ & $16\times16\times256$    \\
Upsampler & $32\times32\times256$    \\
DiffiT ResBlock $\times L_{2}$ & $32\times32\times256$    \\
Upsampler & $64\times64\times256$    \\
DiffiT ResBlock $\times L_{1}$ & $64\times64\times128$    \\
Head & $64\times64\times3$    \\
\bottomrule
\end{tabular}
\end{table}

\subsection{Latent Space}
\begin{figure*}[h]
    \centering
    \includegraphics[width=0.83\textwidth]{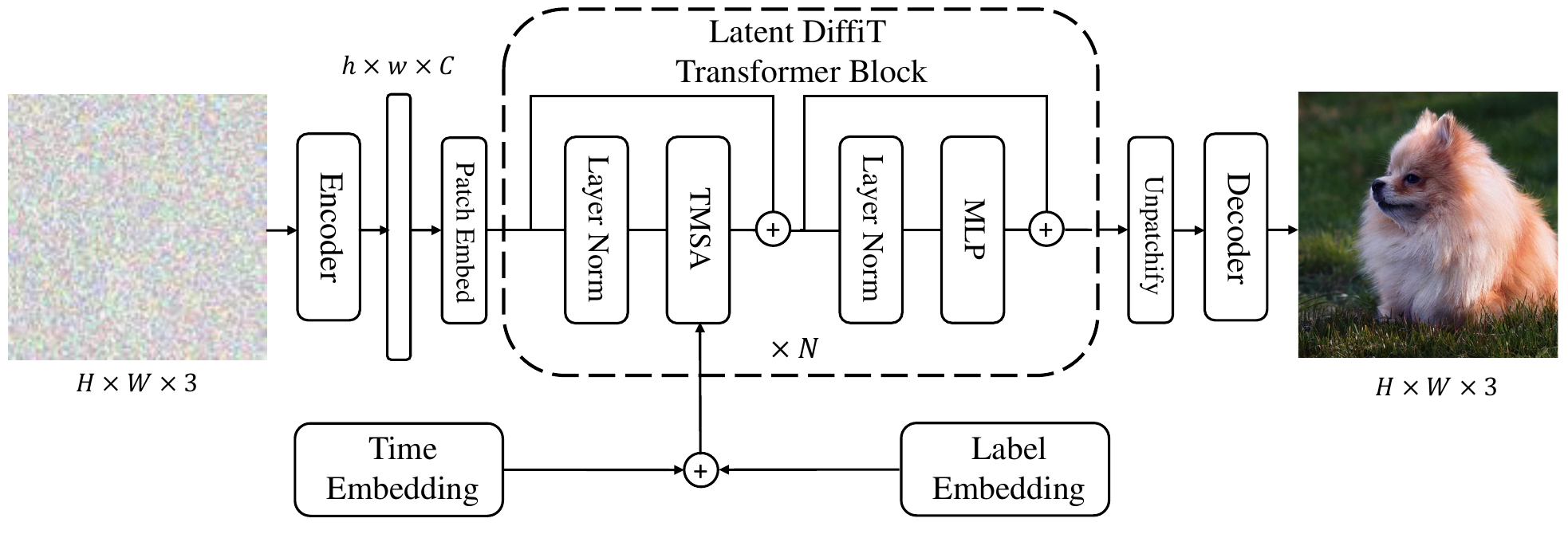}
    \caption{Overview of the latent DiffiT framework.}
    \label{fig:arch-framework-abl}
\end{figure*}

In Fig~\ref{fig:arch-framework-abl}, we illustrate the architecture of the latent DiffiT model. Our model is comparable to DiT-XL/2-G variant which
032 uses a patch size of 2. Specifically, we use a depth of 30 layers with hidden size dimension of 1152, number of heads dimension of 16 and MLP ratio of 4. In addition, for the classifier-free guidance implementation, we only apply the guidance to the first three input channels with a scale of $(1+\mathbf{x})$ where $\mathbf{x}$ is the input latent. 

\section{Implementation Details}
\label{sec:implement}
\subsection{Image Space}
We strictly followed the training configurations and data augmentation strategies of the EDM~\cite{karras2022elucidating} model for the experiments on CIFAR10~\cite{krizhevsky2009learning}, and FFHQ-64~\cite{karras2019style} datasets, all in an unconditional setting. All the experiments were trained for 200000 iterations with Adam optimizer~\cite{kingma2014adam} and used PyTorch framework and 8 NVIDIA A100 GPUs. We used batch sizes of 512 and 256, learning rates of $1 \times 10^{-3}$ and $2\times10^{-4}$ and training images of sizes $32 \times 32$ and $64 \times 64$ on experiments for CIFAR10~\cite{krizhevsky2009learning} and FFHQ-64~\cite{karras2019style} datasets, respectively.

We use the deterministic sampler of EDM~\cite{karras2022elucidating} model with 18, 40 and 40 steps for CIFAR-10 and FFHQ-64 datasets, respectively. For FFHQ-64 dataset, our DiffiT network spans across 4 different stages with 1, 2, 2, 2 blocks at each stage. We also use window-based attention TMSA with local window size of 8 at each stage. For CIFAR-10 dataset, the DiffiT network has 3 stages with 2 blocks at each stage. Similarly, we compute attentions on local windows with size 4 at each stage. Note that for all networks, the resolution is decreased by a factor of 2 in between stages. However, except for when transitioning from the first to second stage, we keep the number of channels constant in the rest of the stages to maintain both the number of parameters and latency in our network. Furthermore, we employ traditional convolutional-based downsampling and upsampling layers for transitioning into lower or higher resolutions. We achieved similar image generation performance by using bilinear interpolation for feature resizing instead of convolution. For fair comparison, in all of our experiments, we used the FID score which is computed on 50K samples and using the training set as the reference set.

\subsection{Latent Space}
We employ learning rates of $3\times 10^{-4}$ and $1\times 10^{-4}$ and batch sizes of 256 and 512 for ImageNet-256 and ImageNet-512 experiments, respectively. We also use the exponential moving average (EMA) of weights using a decay of 0.9999 for both experiments. We also use the same diffusion hyper-parameters as in the ADM~\cite{dhariwal2021diffusion} model. For a fair comparison, we use the DDPM~\cite{ho2020denoising} sampler with 250 steps and report FID-50K for both ImageNet-256 and ImageNet-512 experiments.  

\section{Qualitative Results}
\label{sec:vis}
\begin{figure*}[h]
\centering

\resizebox{0.8\linewidth}{!}{
\begingroup
\renewcommand*{\arraystretch}{0.3}
\begin{tabular}{c}

\includegraphics[]{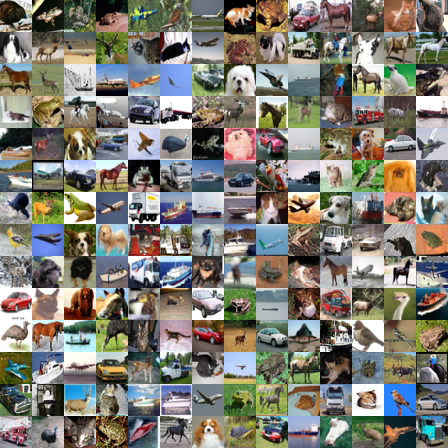} \\
    \vspace{1mm} \\

\end{tabular}
\endgroup
}

\caption{Visualization of uncurated generated images for CIFAR-10~\cite{krizhevsky2009learning} dataset. Best viewed in color.}
\label{fig:cifar-supp}
\end{figure*}

\begin{figure*}[h]
\centering

\resizebox{\linewidth}{!}{
\begingroup
\renewcommand*{\arraystretch}{0.3}
\begin{tabular}{c}

\includegraphics[]{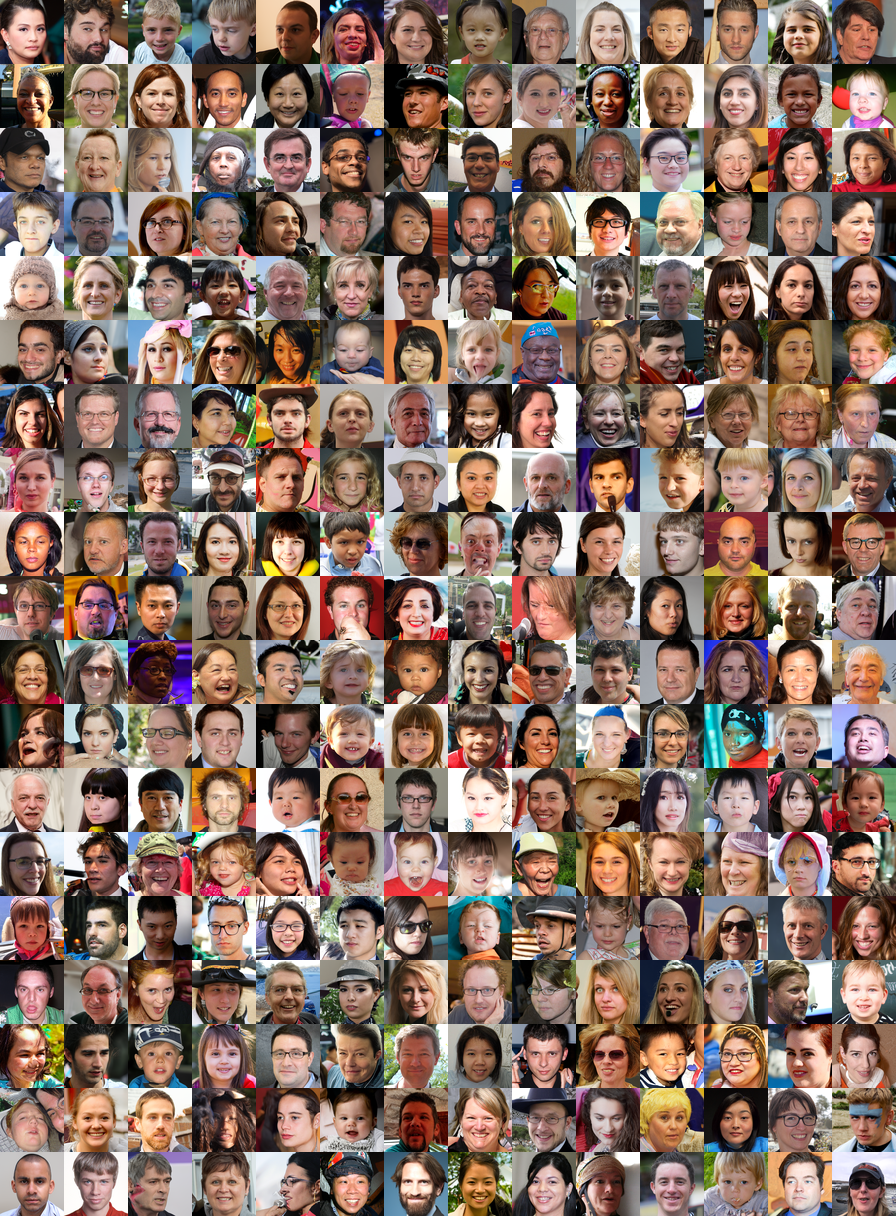} \\
    \vspace{1mm} \\

\end{tabular}
\endgroup
}

\caption{Visualization of uncurated generated images for FFHQ-64~\cite{karras2019style} dataset. Best viewed in color.}
\label{fig:ffhq-supp}
\end{figure*}

\begin{figure*}[h]
\centering
\resizebox{1.\linewidth}{!}{
\begingroup
\renewcommand*{\arraystretch}{0.3}
\begin{tabular}{c}
\includegraphics[width=1\linewidth]{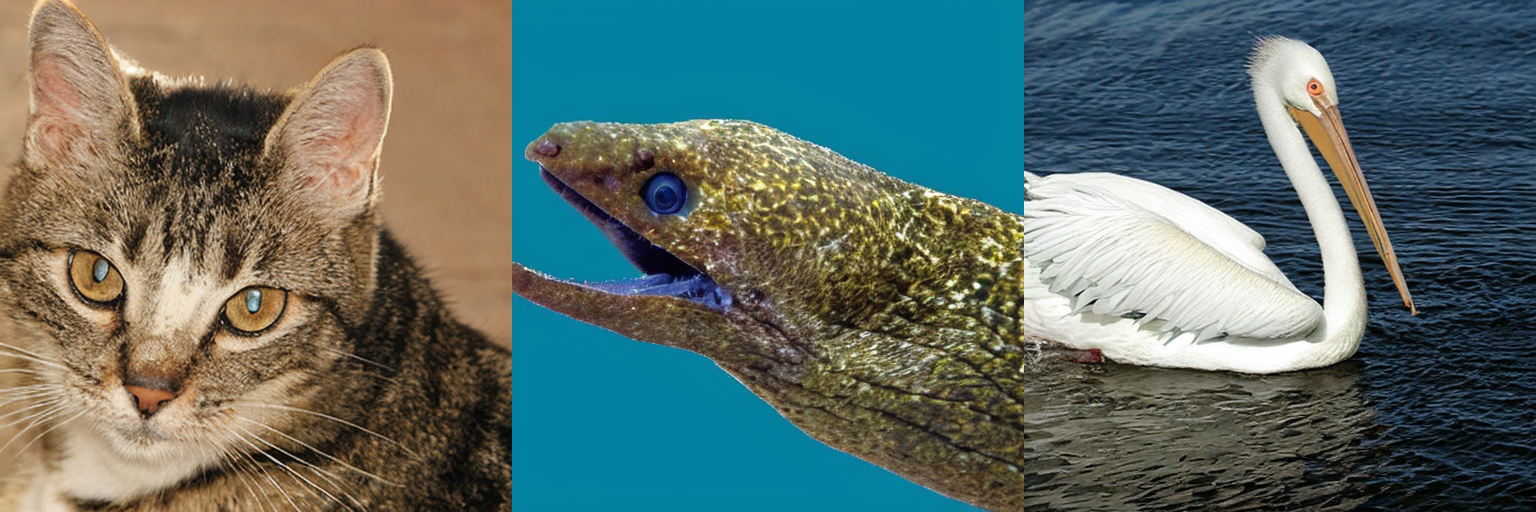} \\ 
\includegraphics[width=1\linewidth]{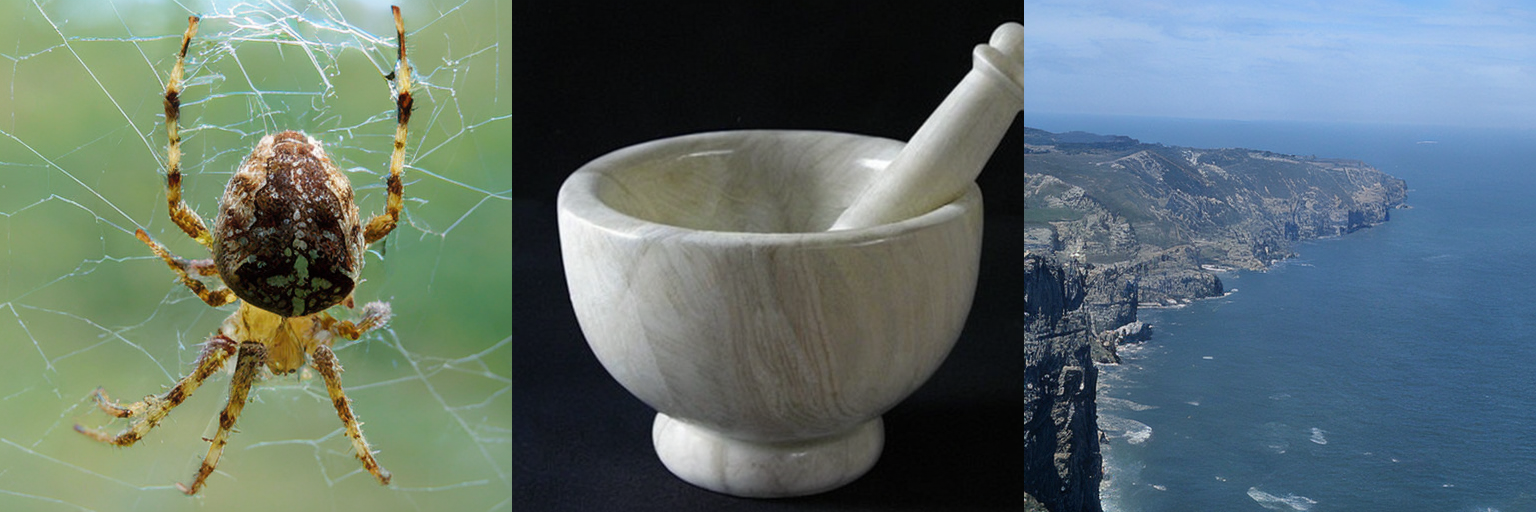} \\ 
\includegraphics[width=1\linewidth]{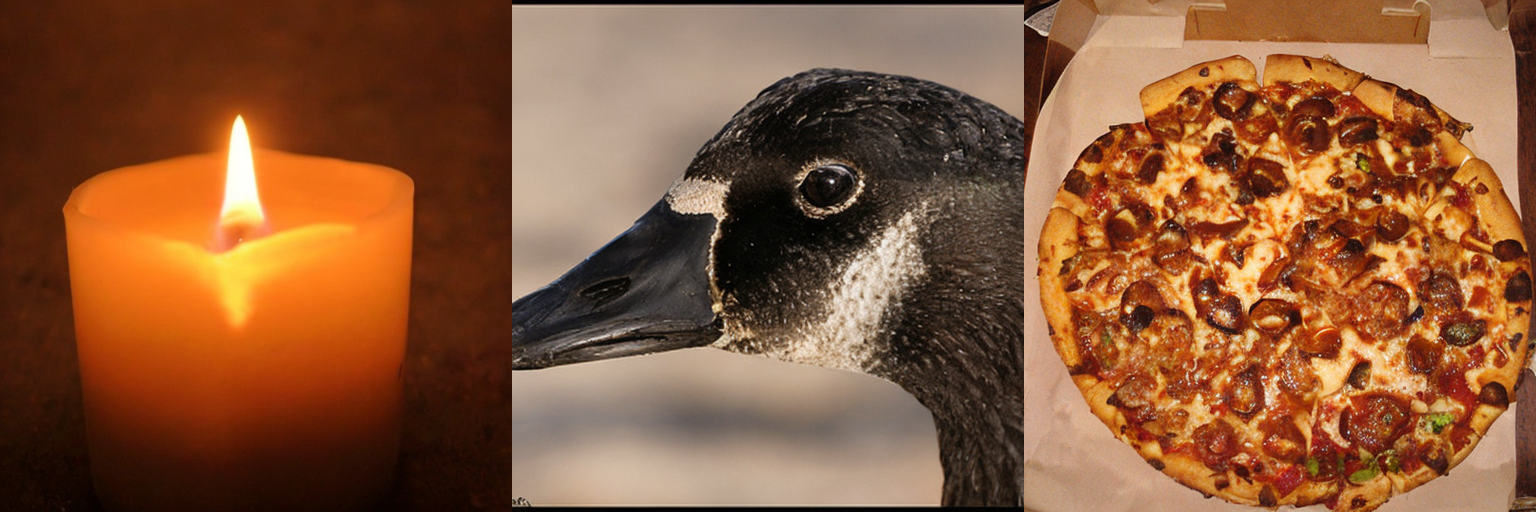} \\
\includegraphics[width=1\linewidth]{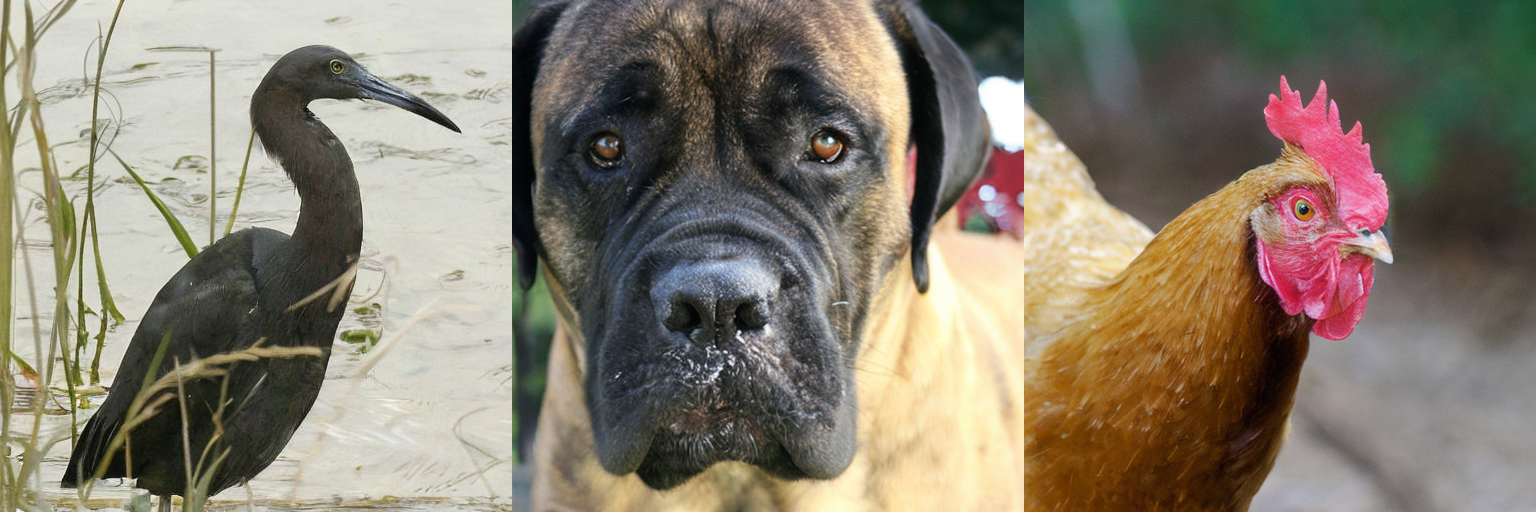} \\
\end{tabular}
\endgroup
}
\caption{Visualization of uncurated generated 512$\times$512 images on ImageNet~\cite{deng2009imagenet} dataset by latent DiffiT model. Images are randomly sampled. Best viewed in color.}
\label{fig:imgnet_512_v11}
\end{figure*}

\begin{figure*}[h]
\centering
\resizebox{1.\linewidth}{!}{
\begingroup
\renewcommand*{\arraystretch}{0.3}
\begin{tabular}{c}
\includegraphics[width=1\linewidth]{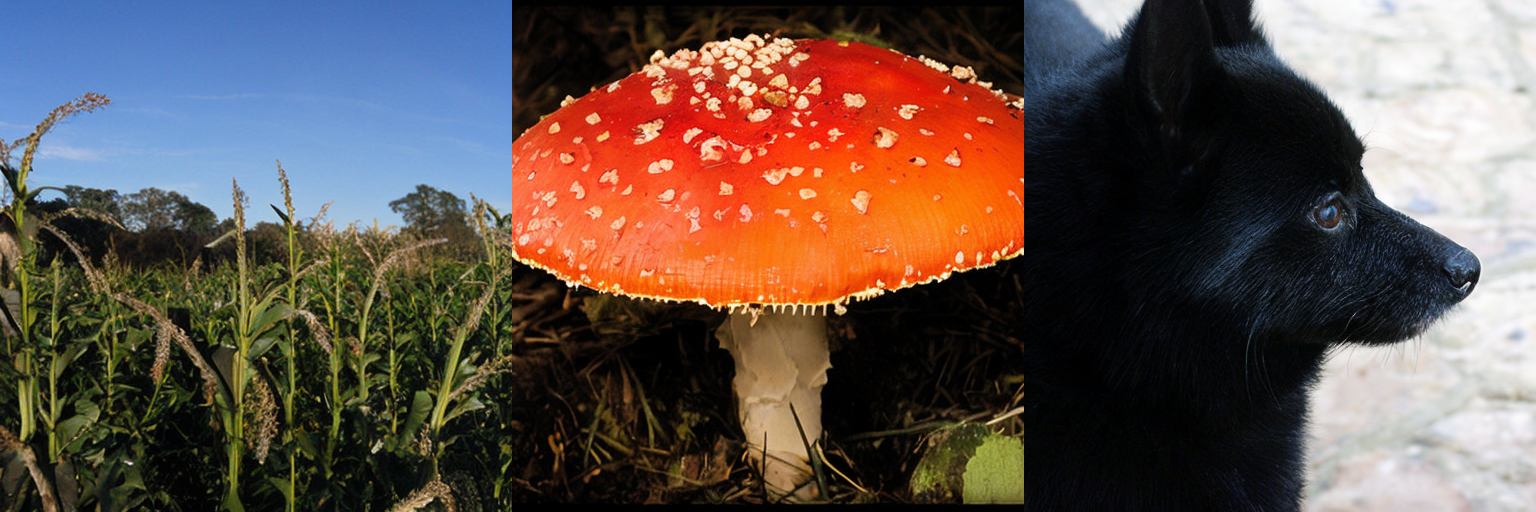} \\ 
\includegraphics[width=1\linewidth]{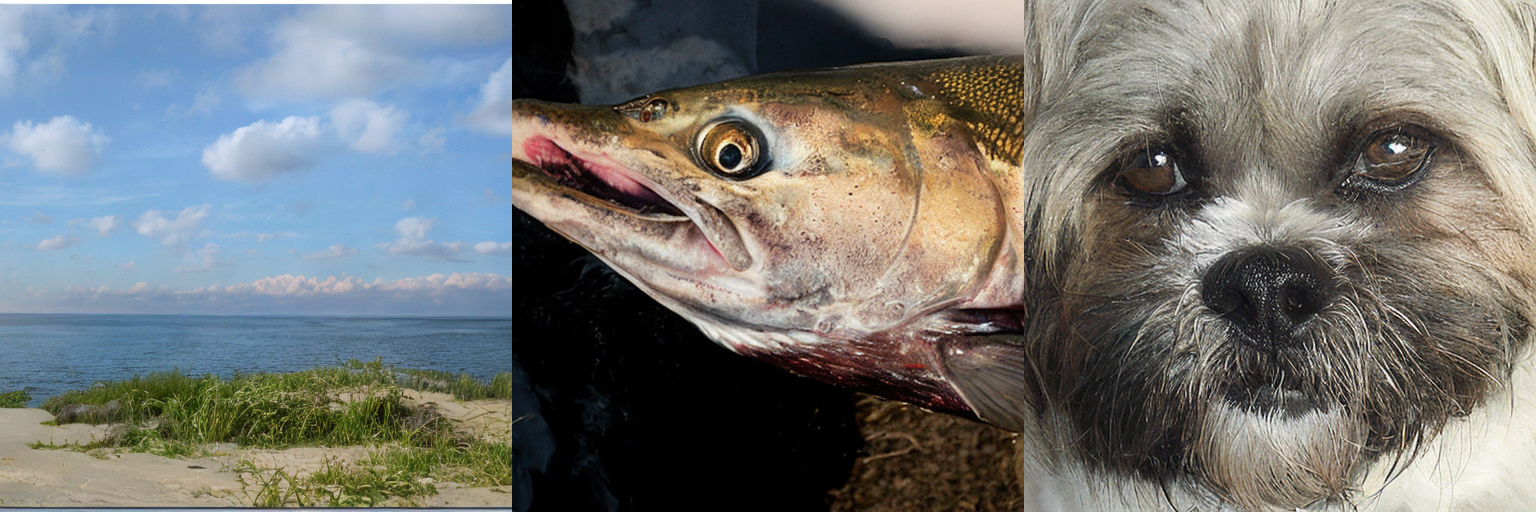} \\ 
\includegraphics[width=1\linewidth]{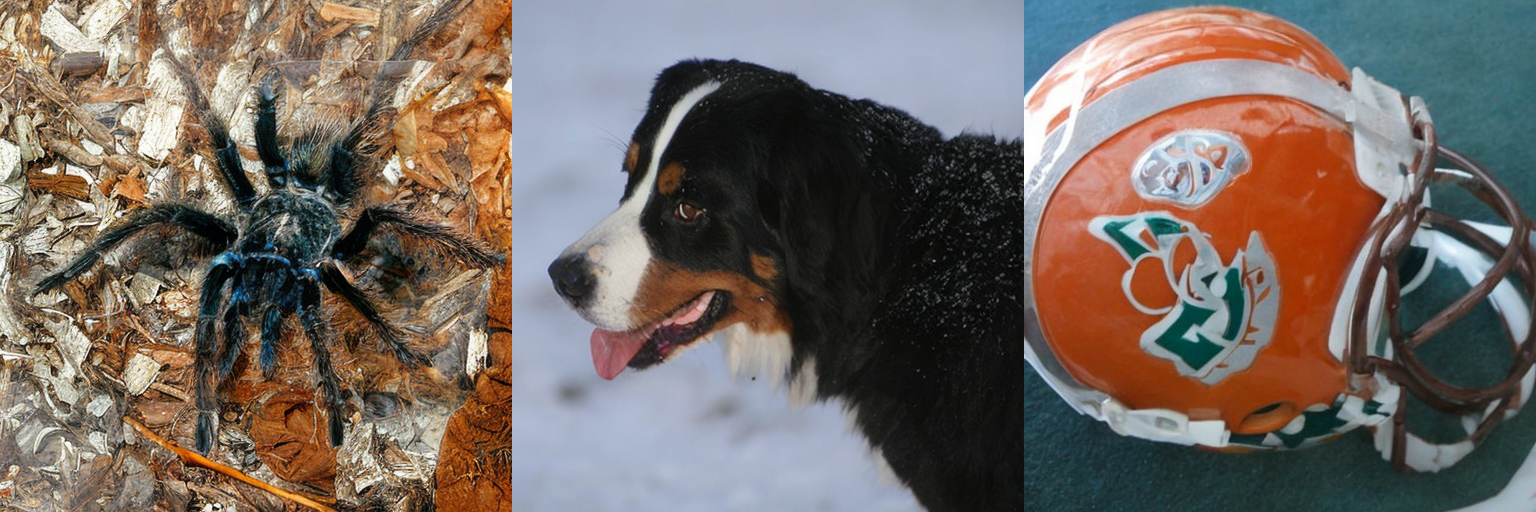} \\
\includegraphics[width=1\linewidth]{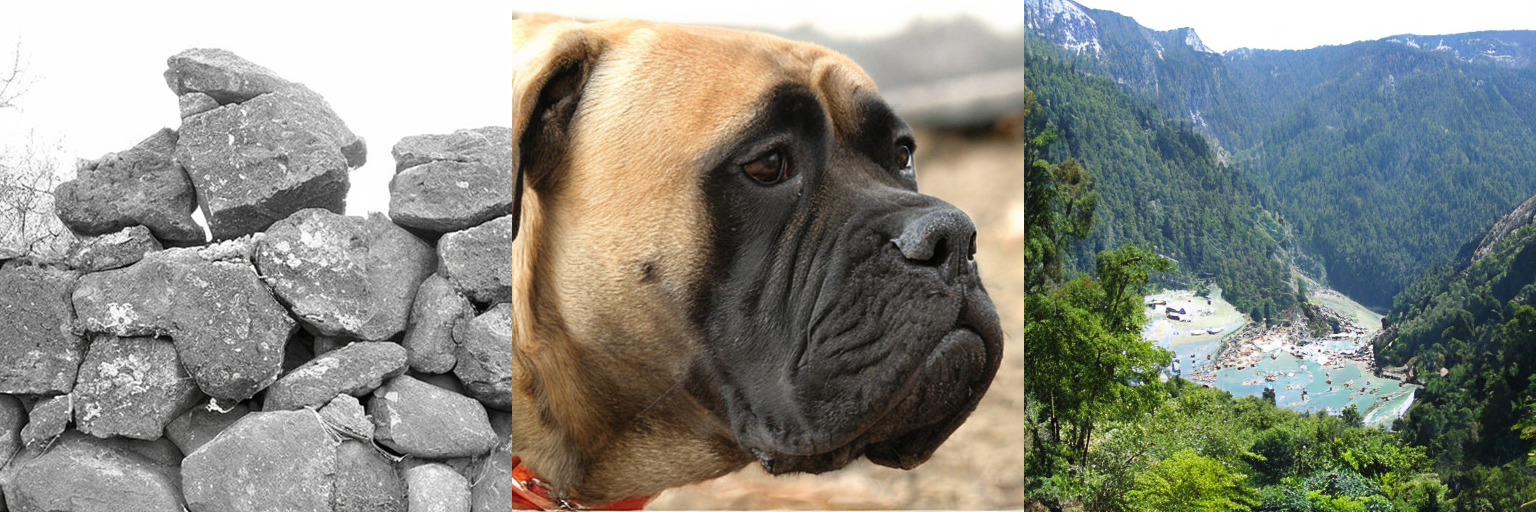} \\
\end{tabular}
\endgroup
}
\caption{Visualization of uncurated generated 512$\times$512 images on ImageNet~\cite{deng2009imagenet} dataset by latent DiffiT model. Images are randomly sampled. Best viewed in color.}
\label{fig:imgnet_512_v12}
\end{figure*}

\begin{figure*}[h]
\centering
\resizebox{1.\linewidth}{!}{
\begingroup
\renewcommand*{\arraystretch}{0.3}
\begin{tabular}{c}
\includegraphics[width=1\linewidth]{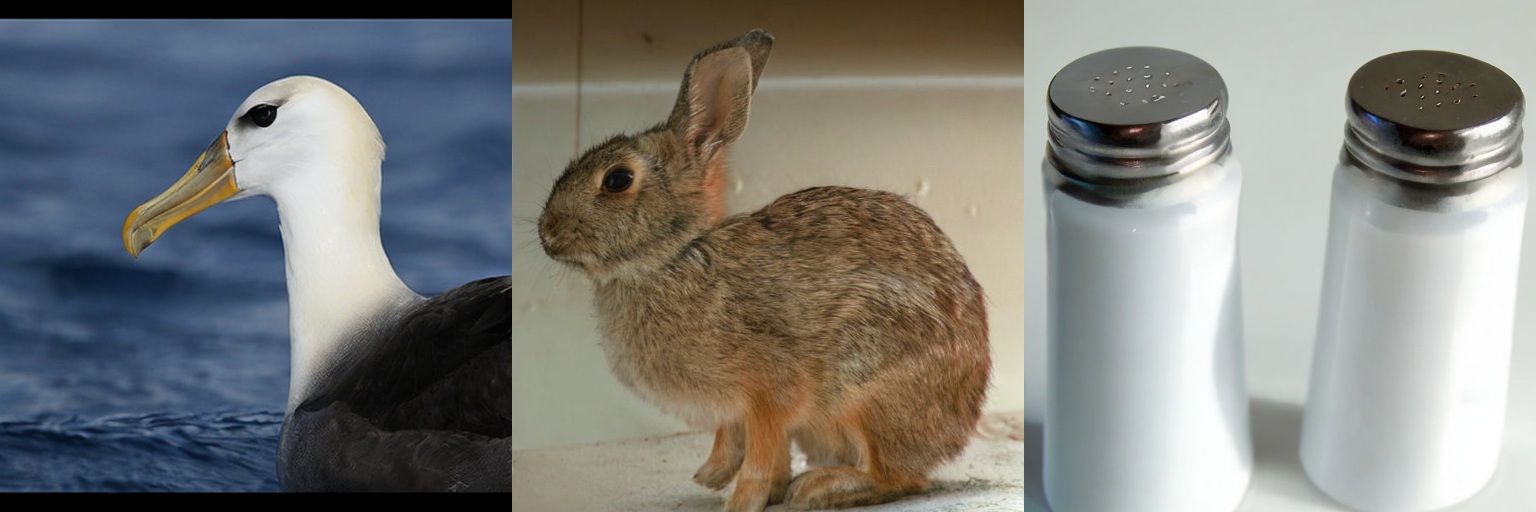} \\ 
\includegraphics[width=1\linewidth]{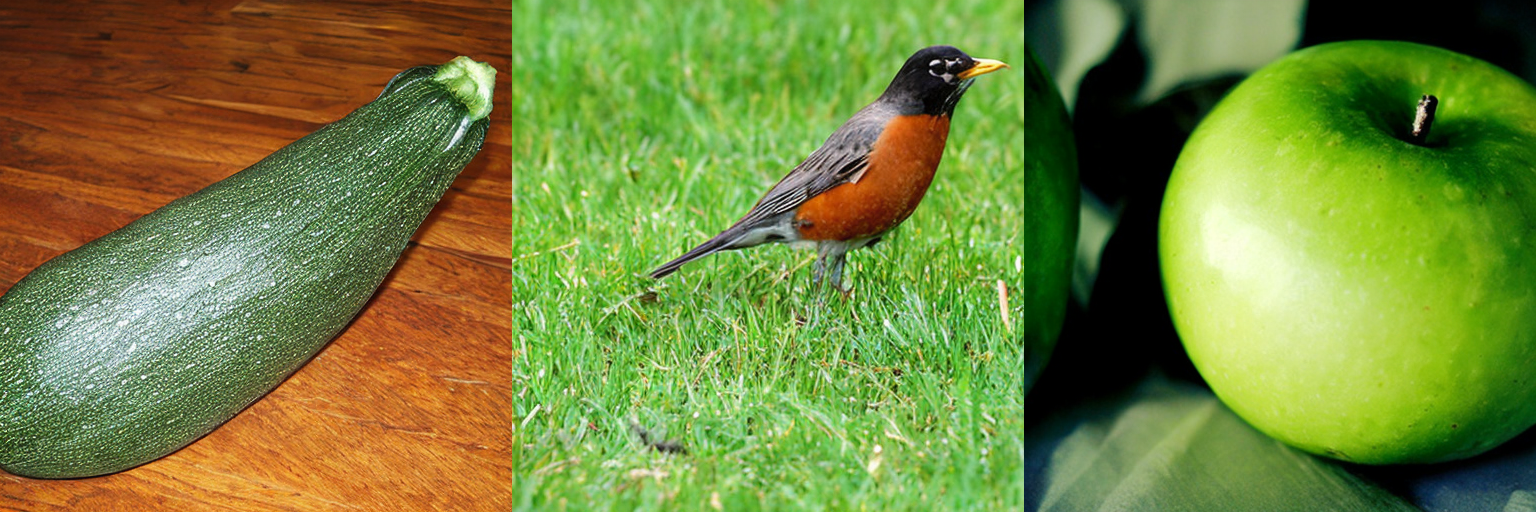} \\ 
\includegraphics[width=1\linewidth]{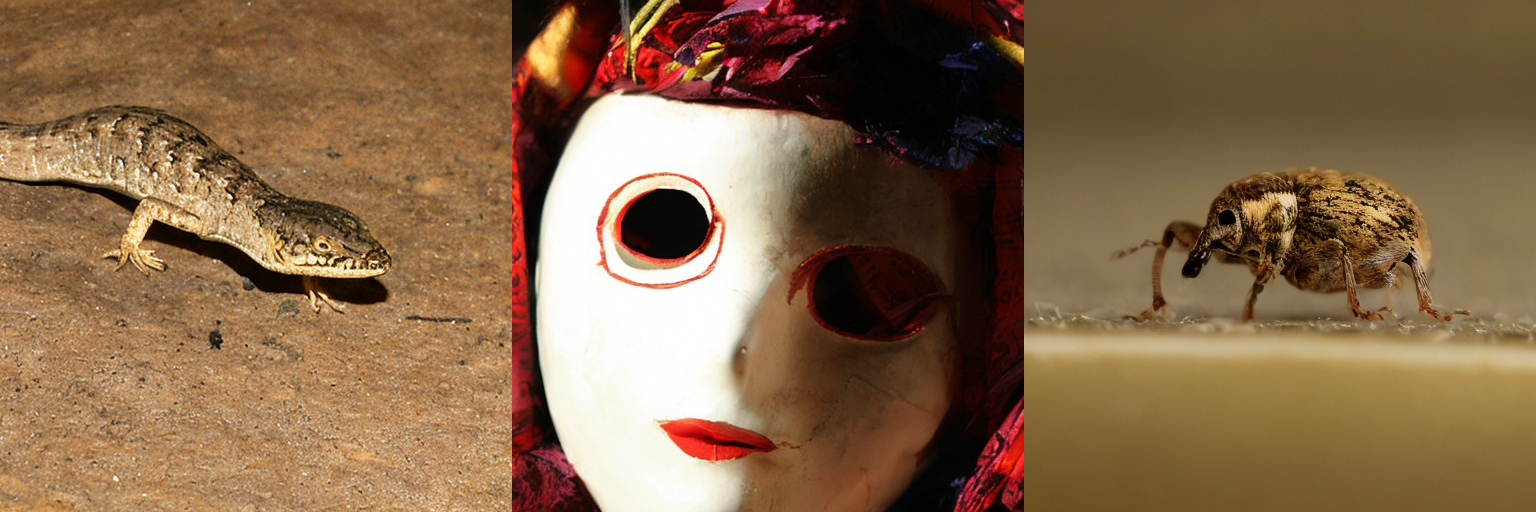} \\
\includegraphics[width=1\linewidth]{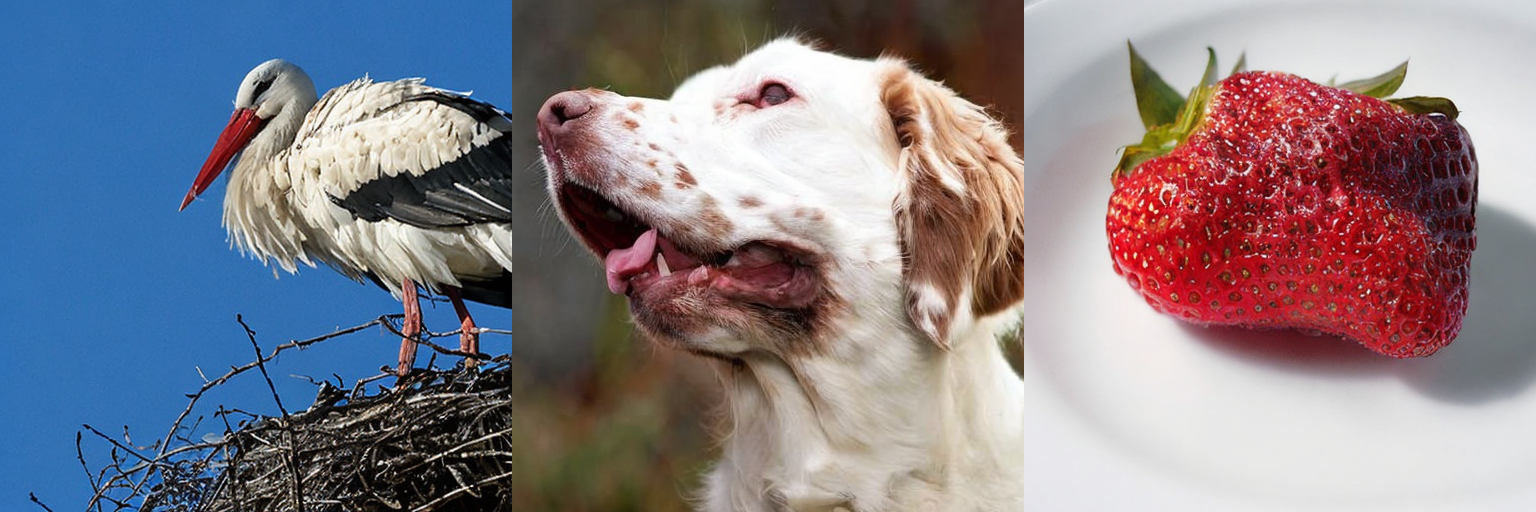} \\
\end{tabular}
\endgroup
}
\caption{Visualization of uncurated generated 512$\times$512 images on ImageNet~\cite{deng2009imagenet} dataset by latent DiffiT model. Images are randomly sampled. Best viewed in color.}
\label{fig:imgnet_512_v1}
\end{figure*}

\begin{figure*}[h]
\centering
\resizebox{1.\linewidth}{!}{
\begingroup
\renewcommand*{\arraystretch}{0.3}
\begin{tabular}{c}
\includegraphics[width=1\linewidth]{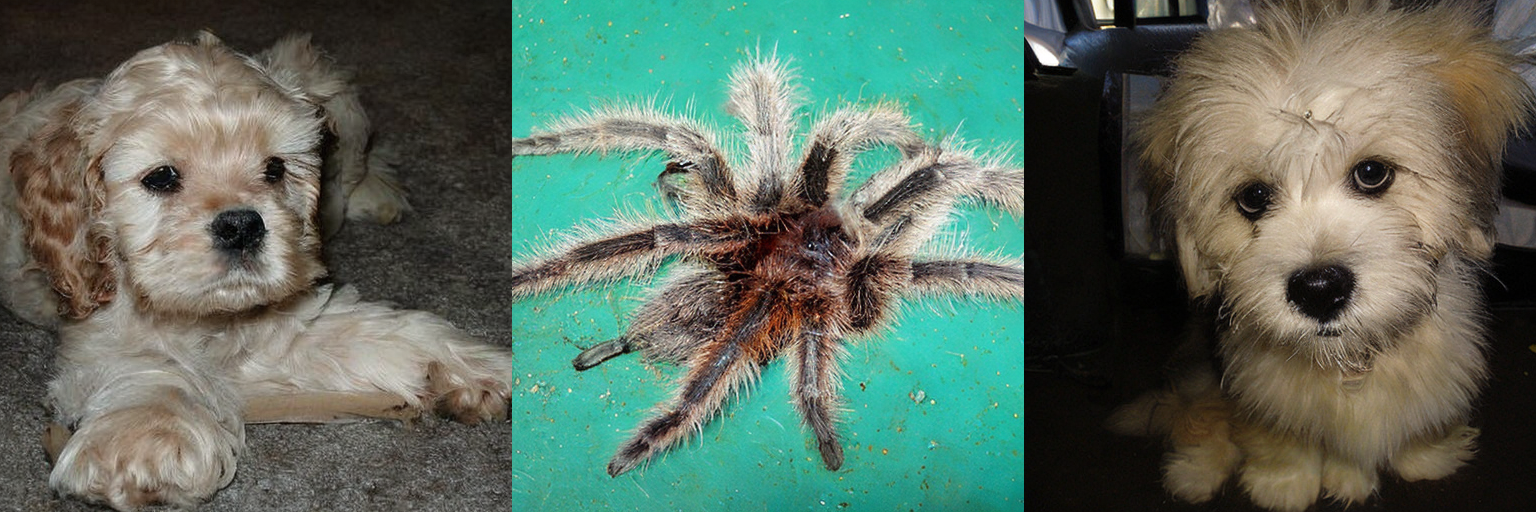} \\ 
\includegraphics[width=1\linewidth]{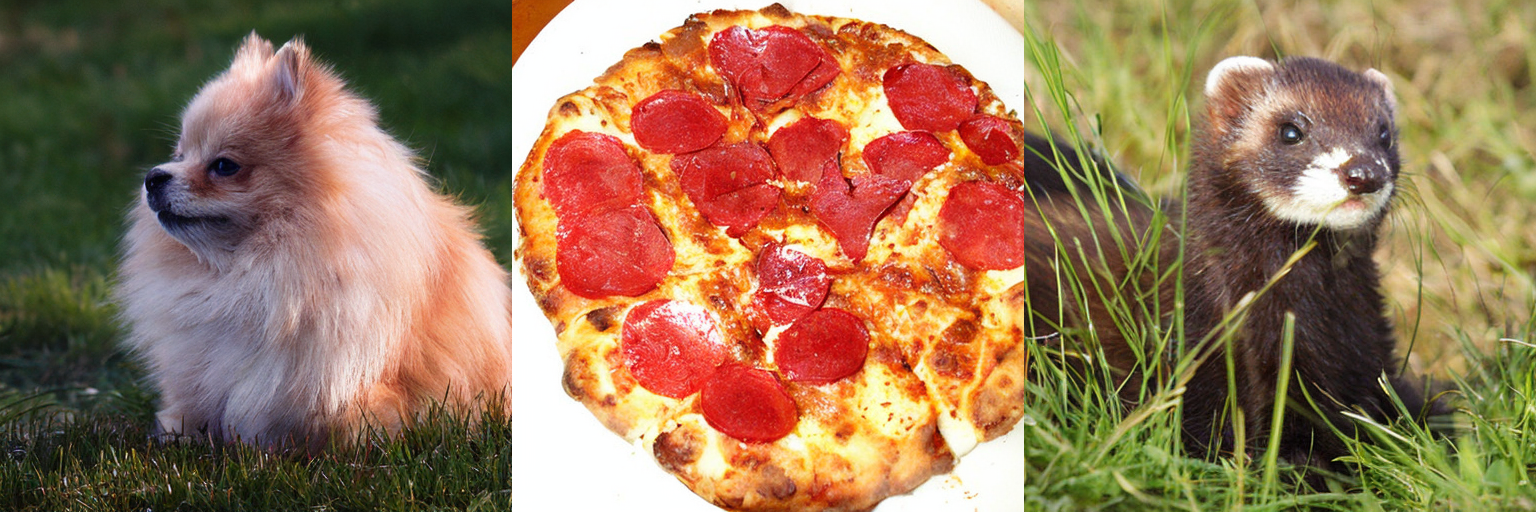} \\ 
\includegraphics[width=1\linewidth]{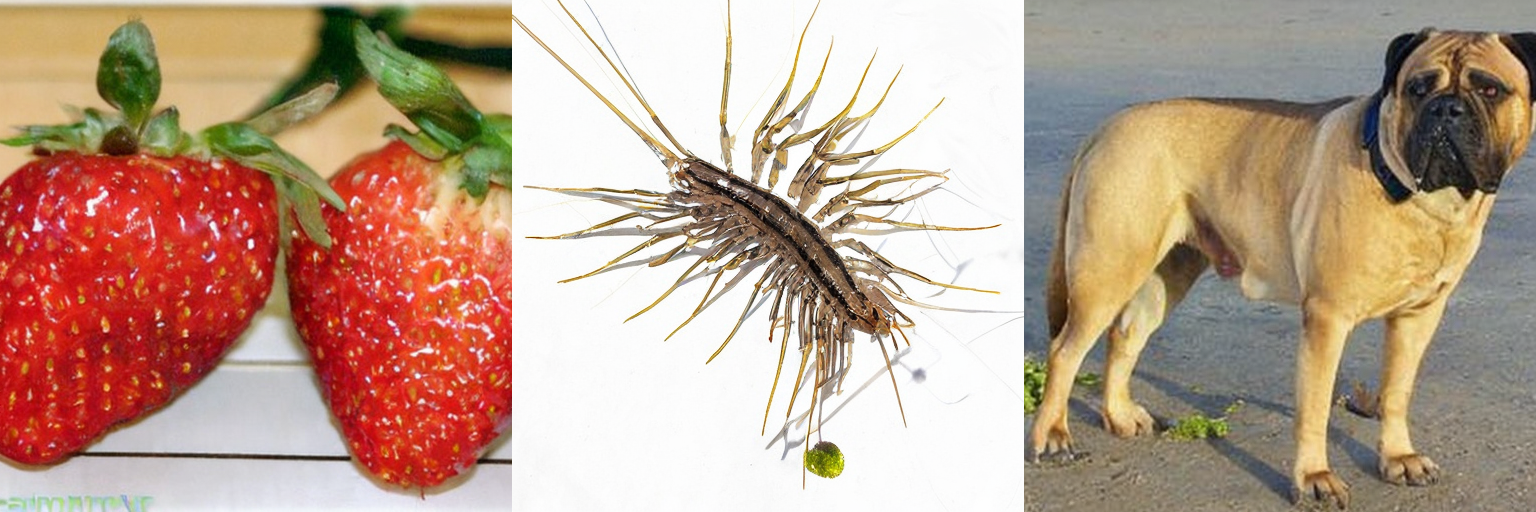} \\
\includegraphics[width=1\linewidth]{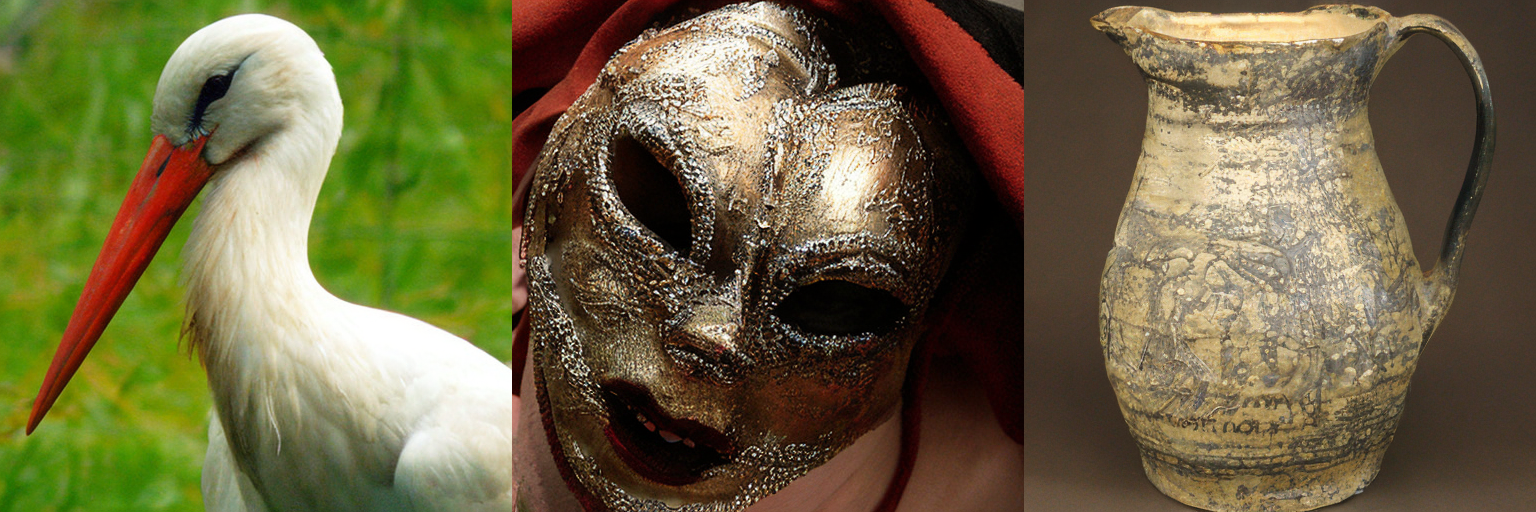} \\
\end{tabular}
\endgroup
}
\caption{Visualization of uncurated generated 512$\times$512 images on ImageNet~\cite{deng2009imagenet} dataset by latent DiffiT model. Images are randomly sampled. Best viewed in color.}
\label{fig:imgnet_512_v2}
\end{figure*}

\begin{figure*}[h]
\centering
\resizebox{1.\linewidth}{!}{
\begingroup
\renewcommand*{\arraystretch}{0.3}
\begin{tabular}{c}
  \includegraphics[width=1\linewidth]{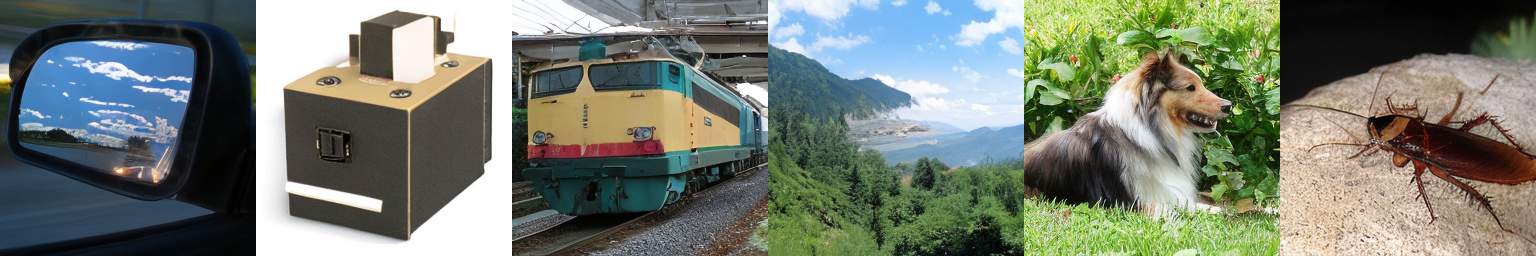} \\ 
\includegraphics[width=1\linewidth]{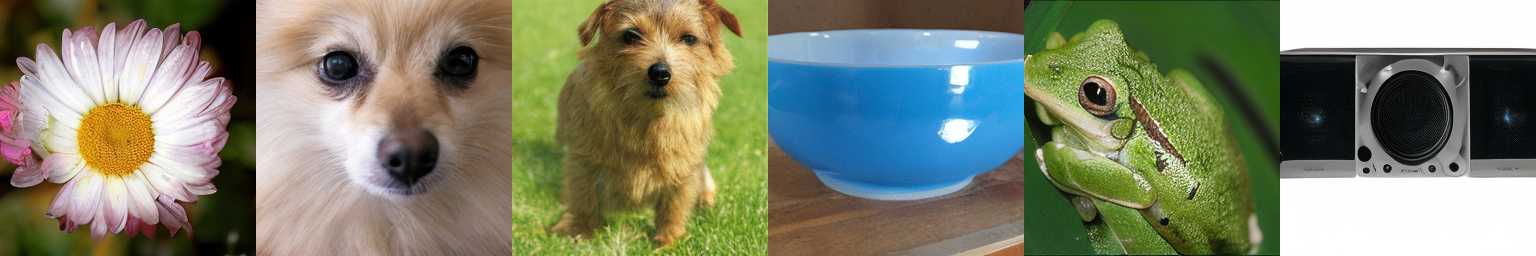} \\
\includegraphics[width=1\linewidth]{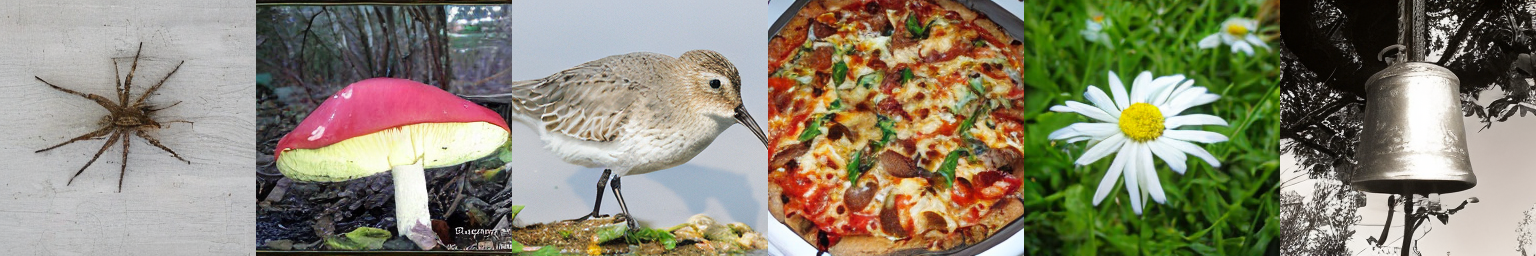} \\
\includegraphics[width=1\linewidth]{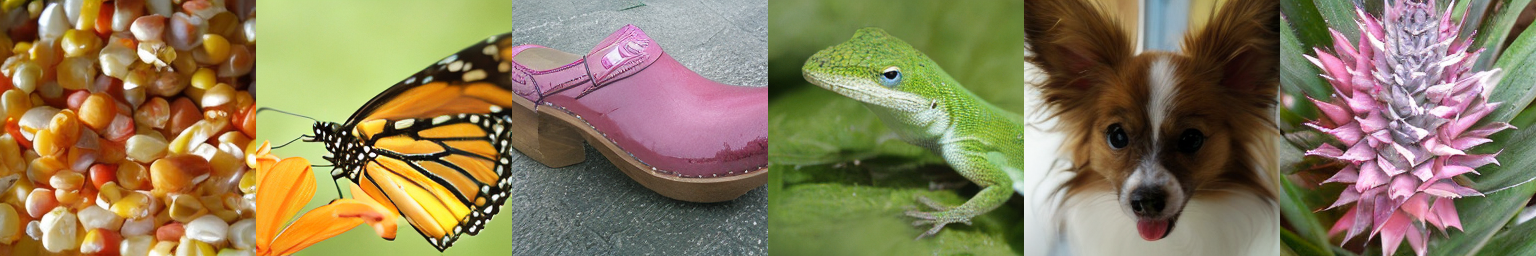} \\
\includegraphics[width=1\linewidth]{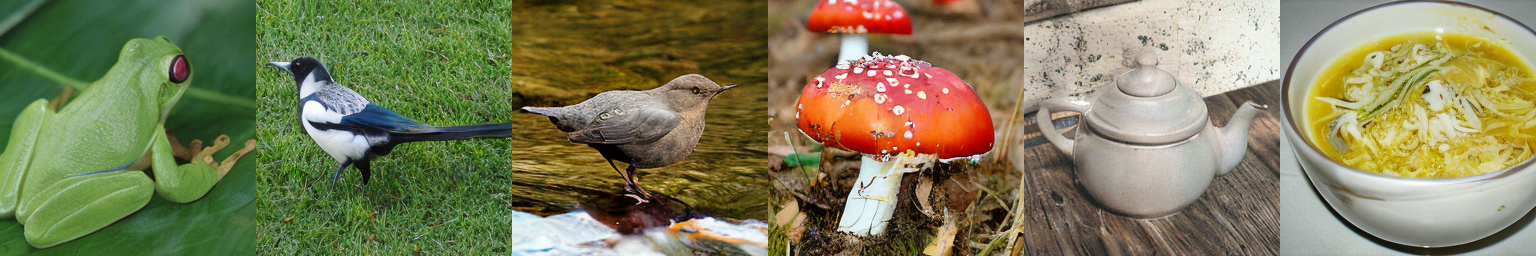} \\
\includegraphics[width=1\linewidth]{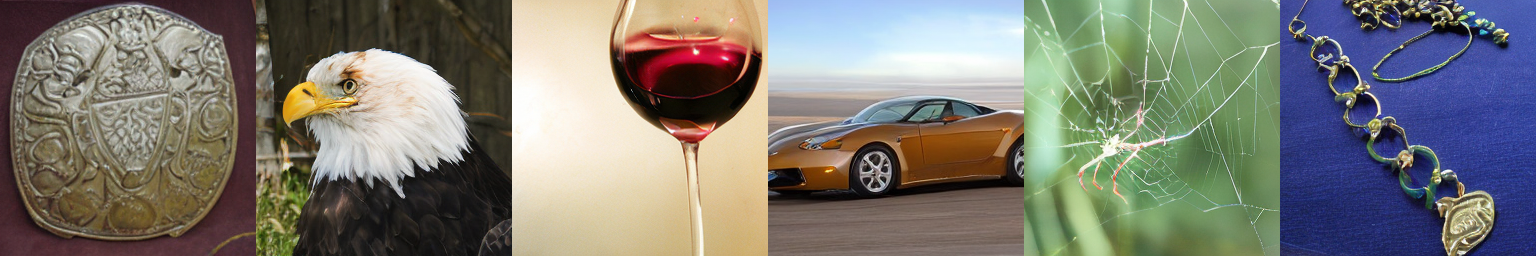} \\
\includegraphics[width=1\linewidth]{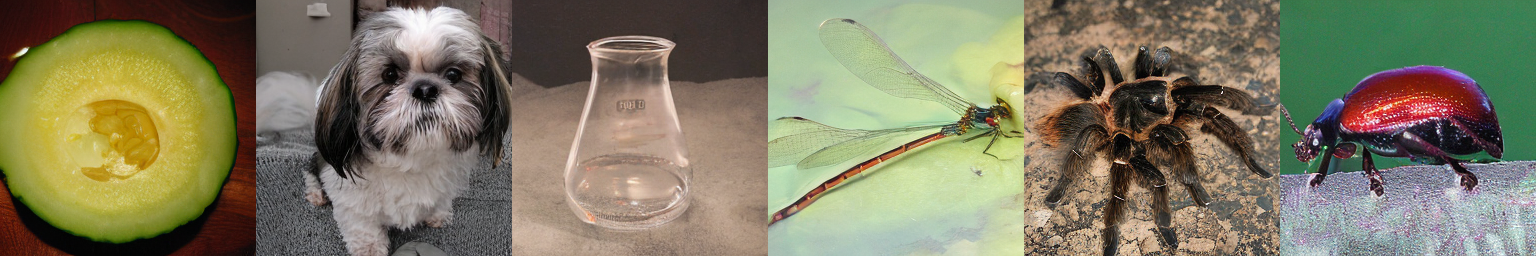} \\
\end{tabular}
\endgroup
}
\caption{Visualization of uncurated generated 256$\times$256 images on ImageNet~\cite{deng2009imagenet} dataset by latent DiffiT model. Images are randomly sampled. Best viewed in color.}
\label{fig:imgnet_256_v1}
\end{figure*}

\begin{figure*}[h]
\centering
\resizebox{1.\linewidth}{!}{
\begingroup
\renewcommand*{\arraystretch}{0.3}
\begin{tabular}{c}
  \includegraphics[width=1\linewidth]{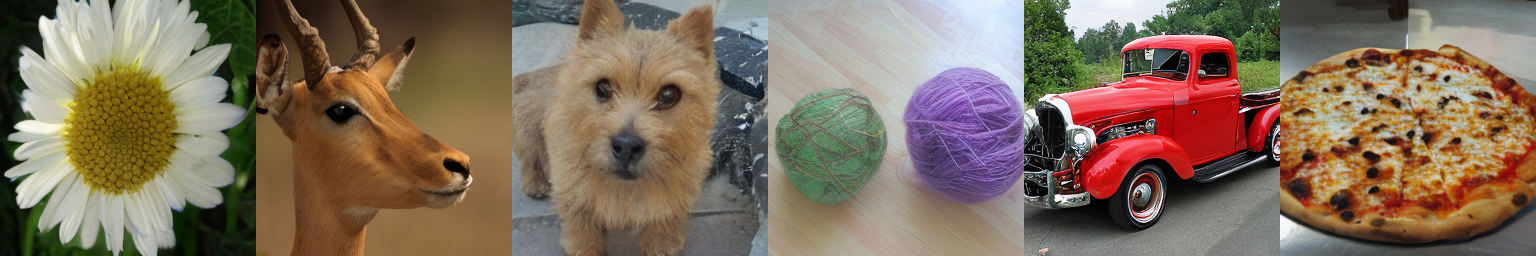} \\ 
\includegraphics[width=1\linewidth]{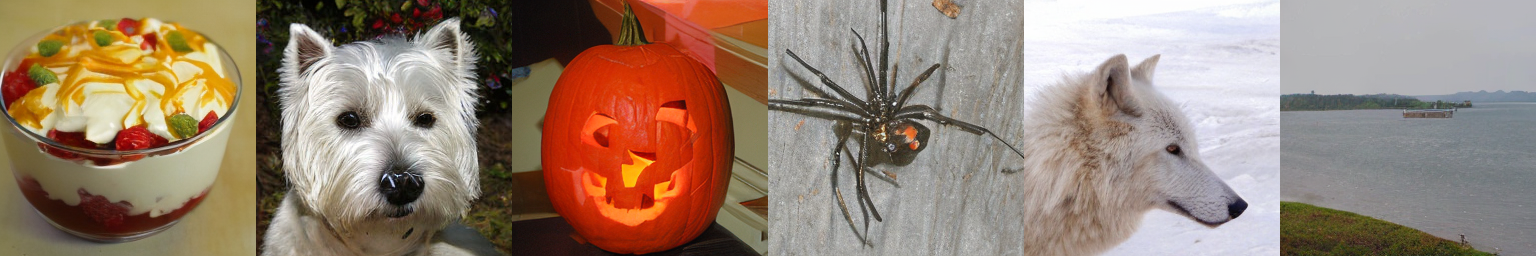} \\
\includegraphics[width=1\linewidth]{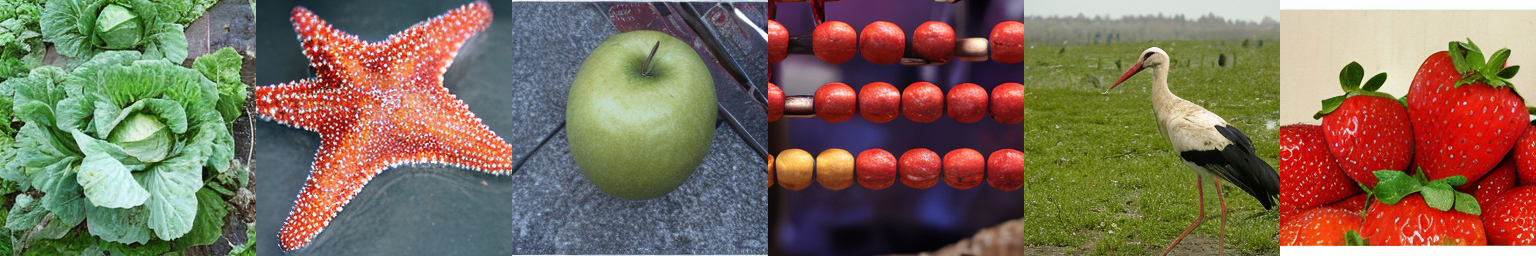} \\
\includegraphics[width=1\linewidth]{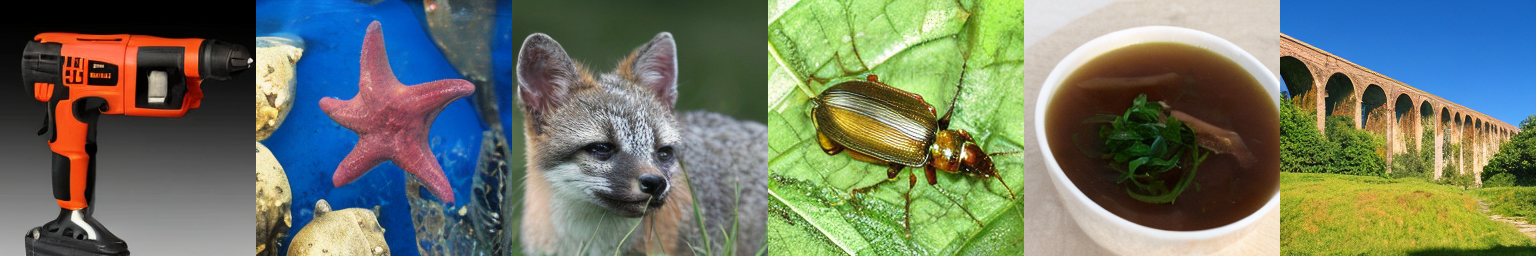} \\
\includegraphics[width=1\linewidth]{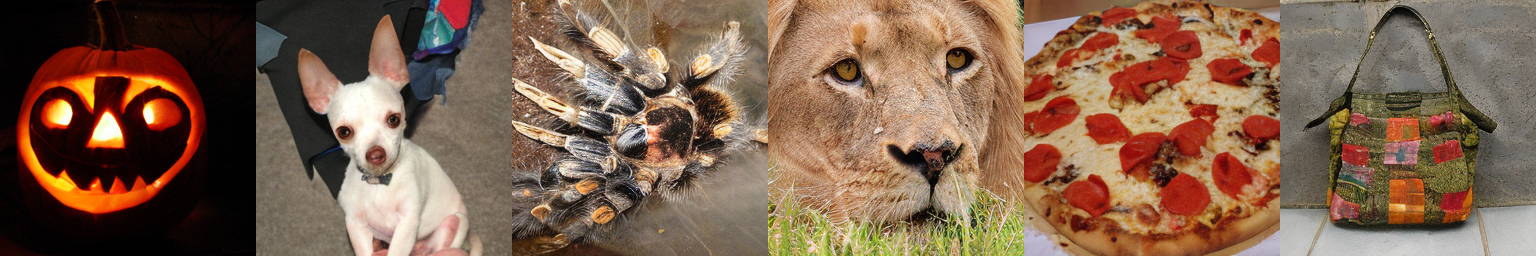} \\
\includegraphics[width=1\linewidth]{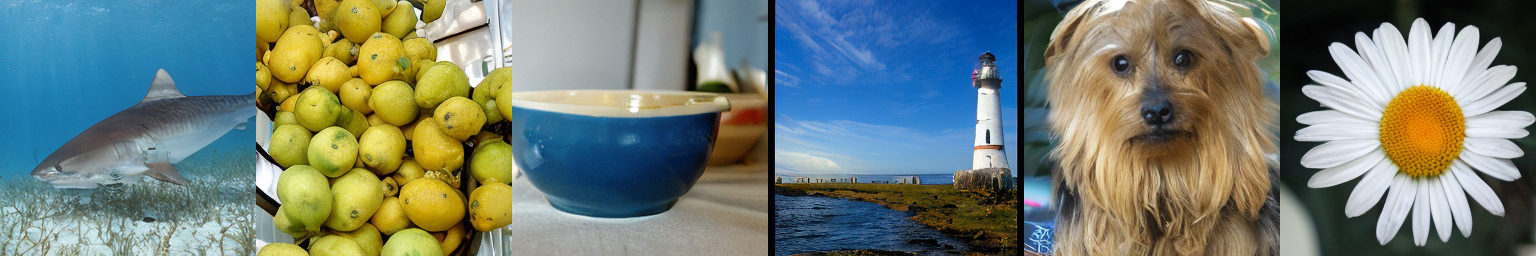} \\
\includegraphics[width=1\linewidth]{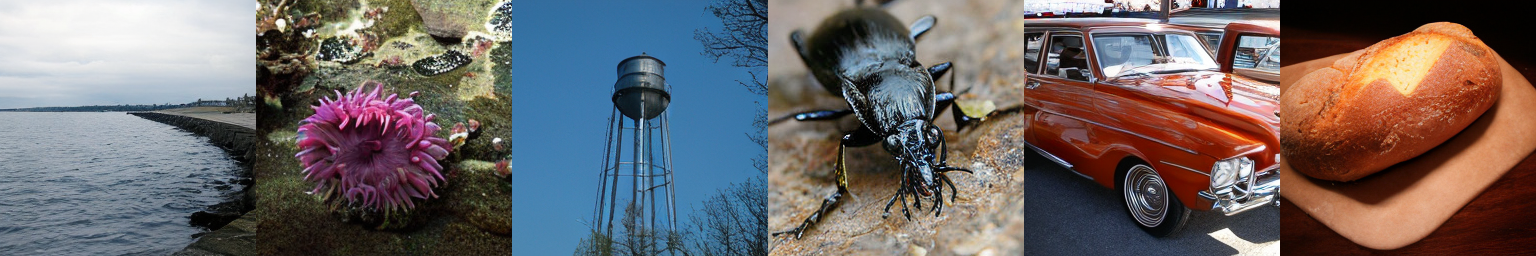} \\
 \\ 
\end{tabular}
\endgroup
}
\caption{Visualization of uncurated generated 256$\times$256 images on ImageNet~\cite{deng2009imagenet} dataset by latent DiffiT model. Images are randomly sampled. Best viewed in color.}
\label{fig:imgnet_256_v2}
\end{figure*}

\begin{figure*}[h]
\centering
\resizebox{1.\linewidth}{!}{
\begingroup
\renewcommand*{\arraystretch}{0.3}
\begin{tabular}{c}
  \includegraphics[width=1\linewidth]{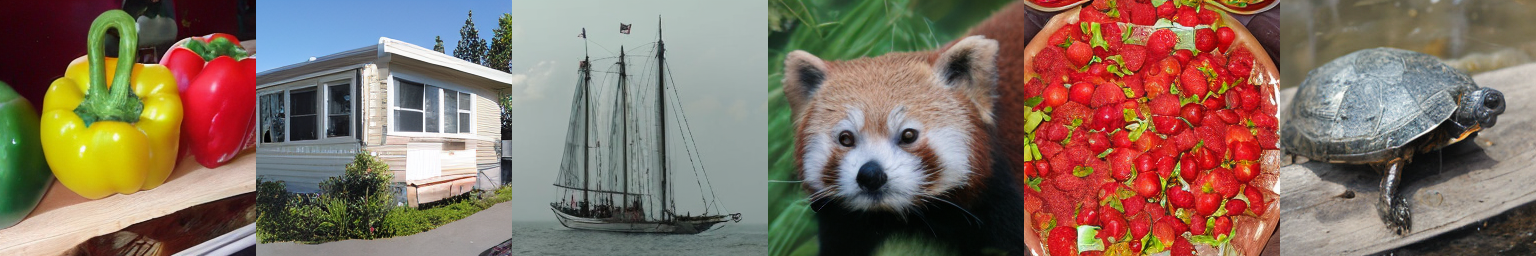} \\ 
\includegraphics[width=1\linewidth]{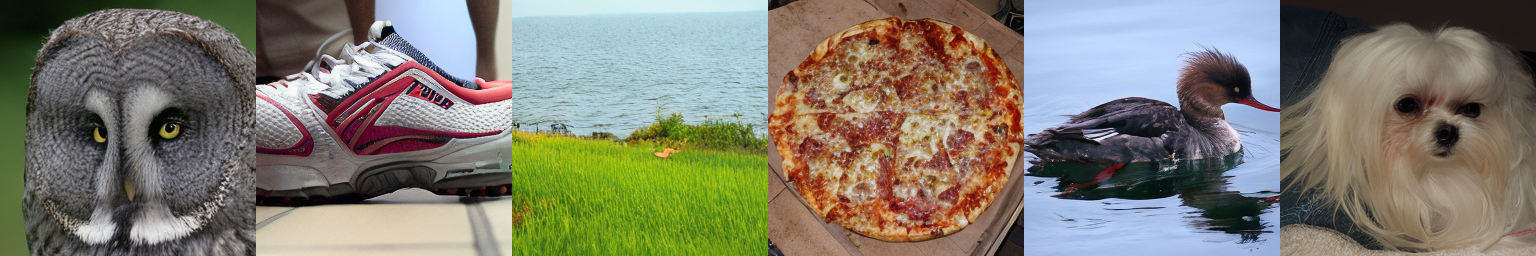} \\
\includegraphics[width=1\linewidth]{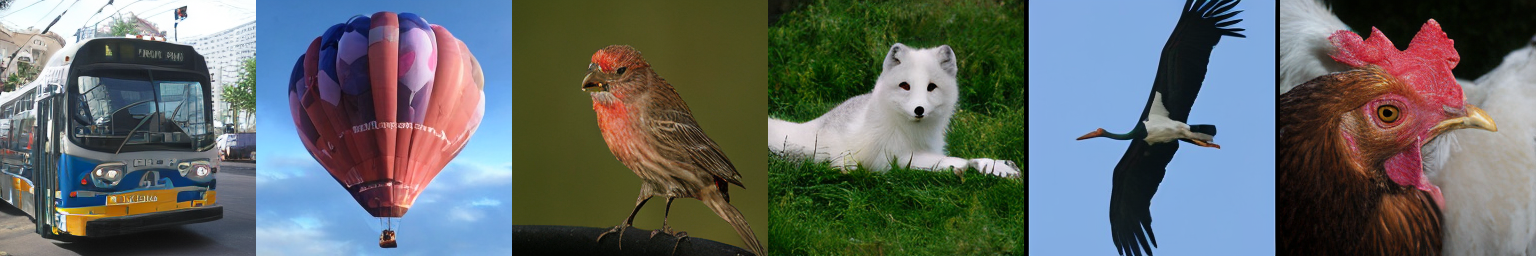} \\
\includegraphics[width=1\linewidth]{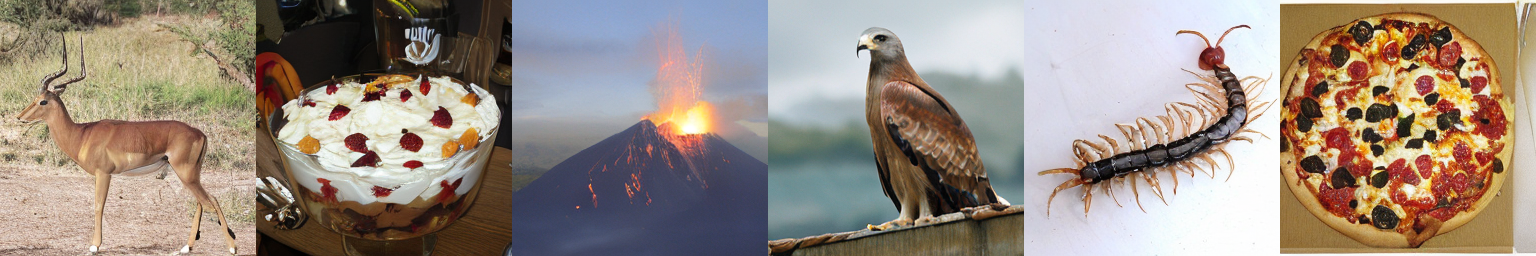} \\
\includegraphics[width=1\linewidth]{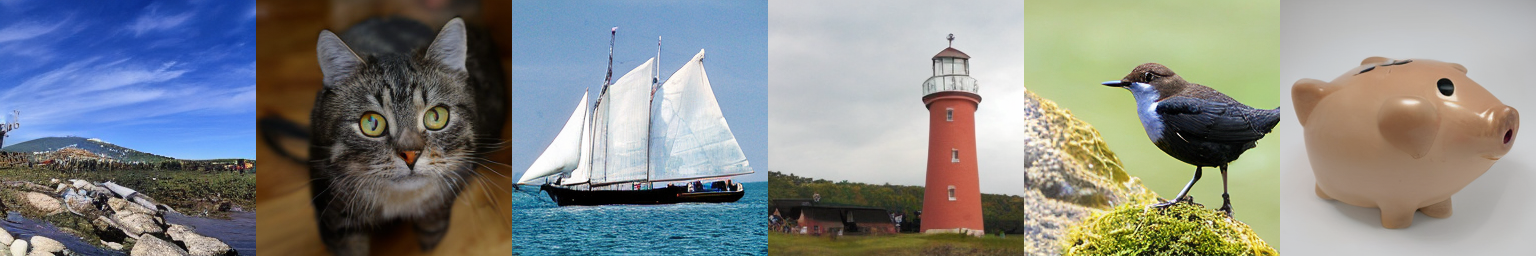} \\
\includegraphics[width=1\linewidth]{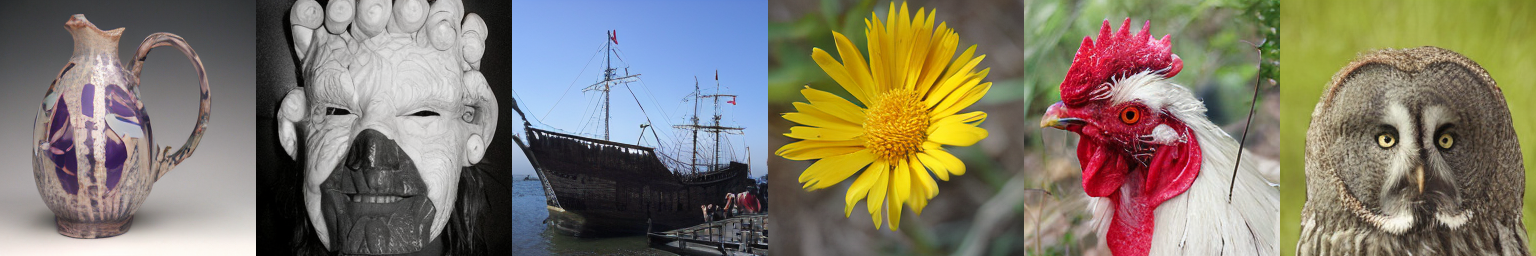} \\
\includegraphics[width=1\linewidth]{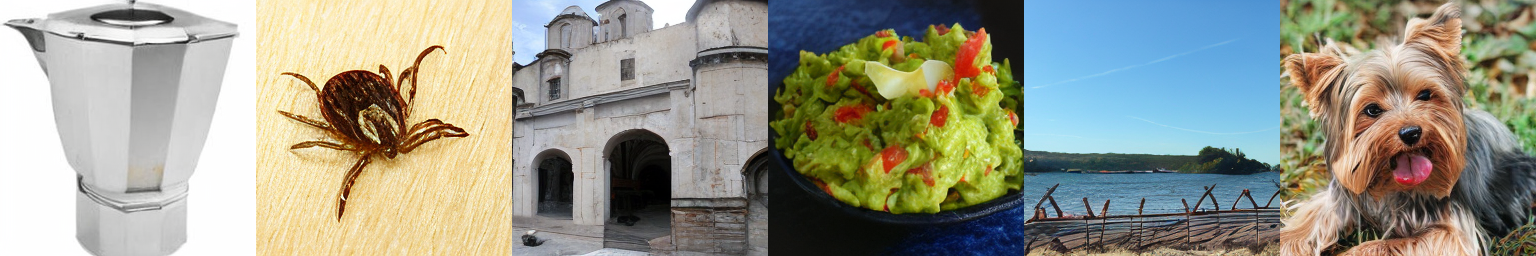} \\
 \\ 
\end{tabular}
\endgroup
}
\caption{Visualization of uncurated generated 256$\times$256 images on ImageNet~\cite{deng2009imagenet} dataset by latent DiffiT model. Images are randomly sampled. Best viewed in color.}
\label{fig:imgnet_256_v3}
\end{figure*}

We illustrate visualization of generated images for CIFAR-10~\cite{krizhevsky2009learning} and FFHQ-64~\cite{karras2019style} datasets in Figures~\ref{fig:cifar-supp} and \ref{fig:ffhq-supp}, respectively. In addition, in Figures~\ref{fig:imgnet_512_v11}, ~\ref{fig:imgnet_512_v12}, ~\ref{fig:imgnet_512_v1} and ~\ref{fig:imgnet_512_v2}, we visualize the the generated images by the latent DiffiT model for ImageNet-512~\cite{deng2009imagenet} dataset. Similarly, the generated images for ImageNet-256~\cite{deng2009imagenet} are shown in Figures ~\ref{fig:imgnet_256_v1}, \ref{fig:imgnet_256_v2} and \ref{fig:imgnet_256_v3}. We observe that the proposed DiffiT model is capable of capturing fine-grained details and produce high fidelity images across these datasets.       

\end{document}